\documentclass{article}
\usepackage{arxiv}
\usepackage[utf8]{inputenc}
\usepackage[T1]{fontenc}
\usepackage{amsmath,amssymb,amsfonts}
\usepackage{graphicx}
\usepackage{booktabs}
\usepackage{hyperref}
\usepackage{xcolor}
\usepackage{microtype}
\usepackage{natbib}
\usepackage{marvosym}
\newcommand\blfootnote[1]{\begingroup\renewcommand\thefootnote{}\footnote{#1}\addtocounter{footnote}{-1}\endgroup}

\renewcommand{\headeright}{}
\renewcommand{\undertitle}{}

\hypersetup{
  colorlinks=true,
  linkcolor={blue!60!black},
  citecolor={blue!60!black},
  urlcolor={blue!60!black},
}

\title{Measurement noise limits the advantage of nonlinear models over linear models in biomedical prediction}

\author{
  Marc-Andr\'{e} Schulz\textsuperscript{1,2,3,4,\,\Letter}, Kerstin Ritter\textsuperscript{1,2,3,4,5} \\\\
  \small \textsuperscript{1}Hertie Institute for AI in Brain Health, University of T\"{u}bingen, T\"{u}bingen, Germany \\
  \small \textsuperscript{2}T\"{u}bingen AI Center, University of T\"{u}bingen, T\"{u}bingen, Germany \\
  \small \textsuperscript{3}Department of Psychiatry and Neurosciences, Charit\'{e} -- Universit\"{a}tsmedizin Berlin, Berlin, Germany \\
  \small \textsuperscript{4}Bernstein Center for Computational Neuroscience, Berlin, Germany \\
  \small \textsuperscript{5}German Center for Mental Health (DZPG), partner site Tübingen, Germany
}

\date{}

\begin{document}
\maketitle
\blfootnote{\Letter~Corresponding author: \texttt{marc-andre.schulz@uni-tuebingen.de}}

\begin{abstract}
On biomedical tabular data, flexible models such as deep networks, gradient-boosted trees, and kernel methods are repeatedly matched or beaten by linear and logistic regression given the same features. The usual reaction is to treat this as a shortfall to be fixed on the model side, with more data, a better architecture, or more careful tuning, on the assumption that the nonlinear structure is there and the model has failed to capture it. We argue that these fixes cannot help when the binding limit is the measurement rather than the model, as it frequently is in biomedicine. Additive noise blurs the population-optimal predictor, the best prediction achievable with unlimited data, and because blurring removes a function's fine, rapidly varying detail before its broad shape, it erases nonlinear structure faster than linear structure. A degree-$k$ interaction is attenuated by the $k$-th power of feature reliability, while the linear part is attenuated only once. At the reliabilities typical of biomedical measurement, the advantage a flexible model would have over a linear one can vanish even when the underlying biology is strongly nonlinear, and what the noise removes cannot be recovered by a larger cohort or a more flexible model, only by better measurement. On this account the nonlinearity is hidden, not absent, and a tie between linear and flexible models is not by itself a verdict on the biology. The pieces of this argument are classical but sit in fields that rarely meet, regression dilution in epidemiology, reliability theory in psychometrics, and the Ornstein--Uhlenbeck operator in Gaussian analysis. Assembled into an exact excess-risk identity, they offer an answer to a question the machine-learning literature has debated for two decades. Measurement reliability is one of three conditions, alongside sample size and feature representation, that must align for a flexible model to help, and together they leave only a narrow window that most biomedical tasks fall outside. We show across 140 UK Biobank tasks that the gap between flexible and linear models, where it exists, carries the predicted noise signature, and that the three conditions can be separated by intervention but not by a benchmark alone.
\end{abstract}

\section{Why linear models are hard to beat}
\label{sec:intro}

For two decades, the same result has recurred across biomedical machine learning, and the field has never settled on how to explain it. Each time a new kind of data becomes available, structural and functional brain imaging, plasma proteomics, gene expression, electronic health records, polygenic scores, someone eventually compares a flexible model against ordinary linear or logistic regression on the same features, expecting the flexible model to find structure the linear one cannot. It usually does not. Ridge and logistic regression match, and often slightly exceed, deep networks, gradient-boosted trees, and kernel machines, and they do so across model families that share no common inductive bias and at sample sizes well into the tens of thousands \citep{schulz2020different, he2020deep, han2025rethinking, smith2025proteomic, christodoulou2019systematic, kelemen2025performance, hand2006classifier, shwartzziv2021tabular}.

Faced with this, most practitioners reach for one of three explanations, and none of them accounts for the breadth of the pattern. The first explanation is sample size. Flexible models are data-hungry, so perhaps the cohorts are too small. But the parity persists at tens of thousands of subjects on tasks whose feature dimension is small enough that a shortage of data cannot be the explanation \citep{schulz2020different, he2020deep, schulz2024performance}. A second explanation is architecture. Deep networks were designed for images and language, not for tables, so perhaps they carry the wrong inductive bias. Yet kernel methods and gradient-boosted trees, which carry no such mismatch, fail by the same margins on the same problems. A third points to tuning, but careful hyperparameter search does not flip the result in the comparisons cited above. What these explanations have in common is that each assumes the nonlinear structure is present and locates the failure in how it is captured, whether in the model, the architecture, or the data. We think that in many cases the structure is no longer recoverable from the noisy features at all, even though it remains present in the biology. The more productive question is about the prediction problem itself. Under what conditions is the best predictor a flexible model could possibly learn no better than the best linear one, so that the choice of model cannot matter?

The answer we develop is that measurement noise can be the binding constraint, and that recognizing this changes how the common finding should be interpreted. When a linear model consistently wins, the natural conclusion is that the biology is close to linear, that there is little nonlinear structure to find. We argue that this conclusion does not follow. The nonlinear structure may well be present, with the measurement hiding it rather than the biology lacking it. The distinction is not academic, because the two readings prescribe opposite actions. If the relationship really is linear, a practitioner should stop searching for nonlinearity and invest the effort elsewhere. If instead the measurement has hidden a nonlinear relationship, the way to recover it is to measure better, and neither a more flexible model nor a larger cohort will substitute for that.

The reason this explanation has been available for decades without being applied is that its pieces live in fields that rarely read one another. Epidemiologists have long corrected for the way measurement error biases an association toward zero, a phenomenon they call regression dilution \citep{frost2000correcting, macmahon1990blood, hutcheon2010random}. Psychometricians have long distinguished the reliability of a measurement from its validity, and have shown that the reliability of a product of two noisy variables is roughly the product of their reliabilities \citep{spearman1904proof, lordnovick1968, bohrnstedt1978reliability, busemeyer1983analysis, kenny1984estimating}. Gaussian analysis supplies the operator that describes how noise smooths a function mode by mode \citep{odonnell2014analysis, janson1997gaussian}. None of these literatures asks the question a machine-learning practitioner asks, which is whether a flexible model class is worth using at all given the features in hand. Carrying these classical results across to that question is the work of this article. We assemble them into an exact identity for the nonlinear advantage, set measurement alongside sample size and representation as the three conditions a flexible model needs, confirm the predicted signature across 140 UK Biobank tasks, and machine-verify the core results in Lean.

\section{How noise hides nonlinearity}
\label{sec:mechanism}

To see how measurement noise lets a linear model match the best nonlinear one, start from what any model, flexible or not, is trying to estimate. Given the observed features $\tilde X$, the best possible prediction of the target $Y$ is the conditional mean $E[Y \mid \tilde X]$, the function that a model with unlimited data and capacity would converge to. Whether a flexible model can beat a linear one comes down to a single question about this function: is it nonlinear in $\tilde X$? If $E[Y \mid \tilde X]$ is close to linear, then no model, however expressive, has anything to gain, because the linear predictor already captures what is there.

\begin{figure*}[t]
  \centering
  \includegraphics[width=\textwidth]{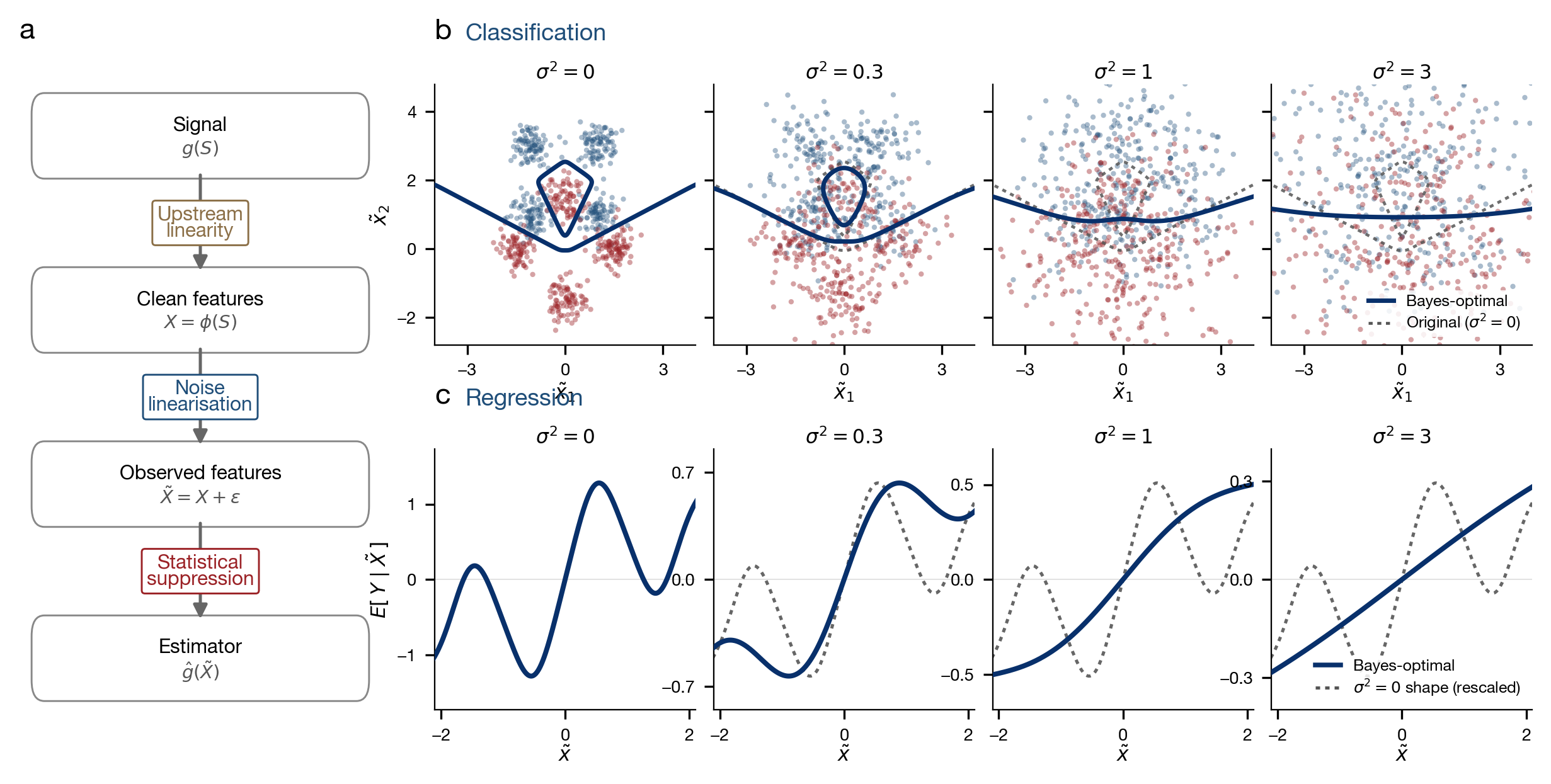}
  \caption{\textbf{Measurement noise straightens the optimal predictor.} (a)~The prediction pipeline runs from biological signal $S$ through feature extraction $X=g(S)$ and measurement $\tilde X = X + \eta$ to a finite-sample estimator. (b)~For classification, the optimal decision boundary at four noise levels, flattening from a curved boundary toward a line. (c)~For regression, the optimal predictor $E[Y\mid\tilde X]$ at four noise levels, straightening toward a linear function as noise grows.}
  \label{fig:framework}
\end{figure*}

Measurement noise enters by reshaping this function. Suppose the clean features $X$ are observed as $\tilde X = X + \eta$ with an independent measurement error $\eta$. The optimal predictor from the noisy features is then a blurred version of the optimal predictor from the clean ones: at each observed value, $E[Y \mid \tilde X]$ averages the clean predictor over the range of clean values that the noisy observation is consistent with. Averaging a function over a neighborhood is smoothing, and smoothing acts unevenly on a function's structure. It leaves the broad, slowly varying parts almost untouched while wearing away the fine, rapidly varying detail. A straight line stays a straight line under smoothing, a gentle bend is rounded off, and a tight oscillation is flattened toward nothing. Figure~\ref{fig:framework} shows the same process for a decision boundary, which loses its curvature and straightens toward a hyperplane, and for a regression function, which loses its bends and straightens toward a line.

The nonlinear content of a prediction function is precisely its fine detail. A linear trend is the broad, slowly varying part of the function; interactions and curvature are the rapid variations, and the higher the order of an interaction, the more rapidly it varies and the finer it is. It follows that as measurement noise grows and the smoothing widens, the optimal predictor loses its nonlinear structure before its linear structure, and within the nonlinear structure it loses the highest-order interactions soonest. The function that survives, the one any model can actually recover from the noisy features, is closer to linear than the clean function was. The relationship in the biology has not changed. The measurement has averaged away its nonlinear part.

One might expect more data to undo the blur, but it cannot. The blurred predictor $E[Y \mid \tilde X]$ is the function that unlimited data converges to, so more subjects sharpen the estimate of it, not the function itself. The blur lifts only when the noise is reduced, through more careful measurement or by averaging repeated measurements of the same subject.

This account can be made exact. Let $\rho = 1/(1+\sigma^2)$ denote the reliability of a feature, the fraction of its observed variance that carries signal for the target, where $\sigma^2$ is the variance of everything else the feature carries, measured relative to that signal. The test-retest reliability, or intraclass correlation (ICC), that biomedical studies report is an upper bound on this $\rho$ rather than equal to it, because the ICC counts every reproducible part of the feature as signal, including reproducible variation unrelated to the target, which for predicting that target acts as noise \citep{spearman1904proof, lordnovick1968}. The two coincide only when all of a feature's reproducible variance is relevant to the outcome. Sorting the clean prediction function by interaction order, through its expansion in the Hermite basis, one finds that the contribution of each order-$k$ component to the excess risk is attenuated by $\rho^k$, the reliability raised to the order \citep{odonnell2014analysis}. Collecting the surviving nonlinear components, the excess error of the best linear predictor over the optimal predictor is
\begin{equation}
  R^*_{\mathrm{lin}} - R_{\mathrm{Bayes}}
  \;=\; \sum_{|\alpha|\ge 2} c_\alpha^2 \prod_{j} \rho_j^{\alpha_j},
  \label{eq:identity}
\end{equation}
where $c_\alpha$ are the coefficients of the clean function and $\alpha$ records its degree in each feature. A pairwise interaction is attenuated by the product of the two reliabilities, a three-way interaction by the product of three, and so on. Each ingredient of this identity is classical. Its degree-one case is regression dilution, the attenuation of a linear coefficient by reliability, long studied in measurement-error statistics \citep{frost2000correcting, carroll2006measurement, fuller1987measurement}. Its product-of-reliabilities form for an interaction is the psychometric result of \citet{bohrnstedt1978reliability} and \citet{busemeyer1983analysis}. The mode-by-mode shrinkage is the Ornstein--Uhlenbeck noise operator of Gaussian analysis \citep{odonnell2014analysis, janson1997gaussian}, and in machine learning, adding noise to a model's inputs is known to act as regularization \citep{bishop1995training, semenova2023path}. The assembled identity adds the model-selection consequence these literatures leave implicit: the entire optimal predictor linearizes, at a rate set by feature reliability raised to the interaction order, so feature reliability, raised to the interaction order, caps the advantage a flexible model can have over a linear one for a given amount of nonlinear structure. Both the identity and its more demanding classification counterpart are machine-verified in the Lean~4 proof assistant (\S\ref{sec:si-lean}).

The single fact to carry forward from Equation~\ref{eq:identity} is the gap it opens between linear and nonlinear structure. Because a linear effect is attenuated by reliability while a $k$-th order interaction is attenuated by reliability to the $k$-th power, the nonlinear advantage that a flexible model could exploit decays faster than the linear signal it would have to beat, and faster still at higher order (Table~\ref{tab:attenuation}). At a feature reliability of $0.9$, near the top of the range for brain-imaging features such as large subcortical volumes, a three-way interaction already retains less than three-quarters of its variance while the linear signal retains nine-tenths. At a reliability of $0.5$, toward the noisier end of biomedical measurement, the linear signal is halved but a three-way interaction is cut to an eighth. This differential attenuation sets a ceiling on the nonlinear advantage. For a given amount of nonlinear structure, that ceiling depends on feature reliability and the order of the interaction, and no choice of model and no larger sample can overcome it, because Equation~\ref{eq:identity} concerns the population predictor itself. By the time features are noisy enough, that ceiling is near zero, even though linear prediction from the same features may remain useful.

\begin{table}[ht]
\centering
\caption{\textbf{Fraction of an interaction's variance that survives measurement noise, by interaction order.} Each entry, $\rho^k$ for feature reliability $\rho$ and interaction order $k$, is the fraction of the mode's energy that survives, equal to the square of the $\rho^{k/2}$ factor by which noise attenuates its coefficient. A linear effect ($k=1$) is attenuated in proportion to reliability; nonlinear structure decays faster, and faster still as order increases.}
\label{tab:attenuation}
\begin{tabular}{@{}lccc@{}}
\toprule
Feature reliability $\rho$ & Linear ($k{=}1$) & Pairwise ($k{=}2$) & Three-way ($k{=}3$) \\
\midrule
0.9 & 90\% & 81\% & 73\% \\
0.7 & 70\% & 49\% & 34\% \\
0.5 & 50\% & 25\% & 12\% \\
0.3 & 30\% & 9\% & 3\% \\
\bottomrule
\end{tabular}
\end{table}

\section{When a flexible model helps}
\label{sec:implications}

For a flexible model to beat a linear one, two things must be true at once. The population-optimal predictor on the observed features must be nonlinear, and a finite sample must be able to estimate that nonlinearity. Three conditions decide whether both hold, one at each stage as features are built, measured, and then used to fit a model (Fig.~\ref{fig:framework}). Feature reliability and the representation determine whether the optimal predictor is nonlinear at all; sample size determines whether a nonlinearity that is present can be estimated. A tie between the two model classes therefore has more than one cause. The biology may be linear, the noise may have hidden a real nonlinearity, the representation may have absorbed it, or the sample may be too small to recover it. Of the three conditions, sample size is the one practitioners routinely consider; reliability and the representation are the two they seldom check.

Like measurement noise, the representation can flatten the optimal predictor, but it does so before any model is fit. A kernel method earns its power by mapping the inputs into a space where a linear rule suffices, and that map is the whole of its advantage \citep{cover1965geometrical, rahimi2007random}. Biomedical pipelines often build such a map in advance, by hand or by convention, so the model that would have supplied it finds the work done. An engineered composite does this, as when creatinine, age, and sex are folded into estimated GFR through a nonlinear formula, so that a linear term captures a relationship nonlinear in the raw measurements. Aggregation does it too, when averaging many sharply varying parts into one summary measure smooths a response that no line would fit at the finer scale, as when many voxels are pooled into a regional volume or many peptides into a protein level \citep{alfaroalmagro2020confound, sun2023plasma}. A learned encoder reaches the same end by fitting the map from data instead of designing it, which is why a simple linear model on a pretrained encoder's features can rival training the whole network \citep{kumar2022finetuning}. Once the features lie in such a space, the best predictor on them is close to linear, and no flexible model, the kernel included, has anything to add.

Even when the optimal predictor on the features is nonlinear, the sample can be too small to estimate it. A flexible model can take on many shapes, and a small sample does not pin down which one, so many curved fits run close to the training points but disagree elsewhere. A linear model can take few shapes and changes little from one sample to the next, so it carries more bias but much less variance, and on a small cohort that tradeoff favors it even when the truth is nonlinear. The sample needed before the flexible model pulls ahead grows with the number of features and the order of the interactions, and falls with the strength of the signal; biomedical tasks tend to pair high dimension with weak signal, which pushes that point beyond the sizes most cohorts reach. The structure is then present and well measured yet still unrecovered, this time for want of data \citep{hastie2009elements, marek2022reproducible}.

\begin{figure}[t]
  \centering
  \includegraphics[width=0.9\columnwidth]{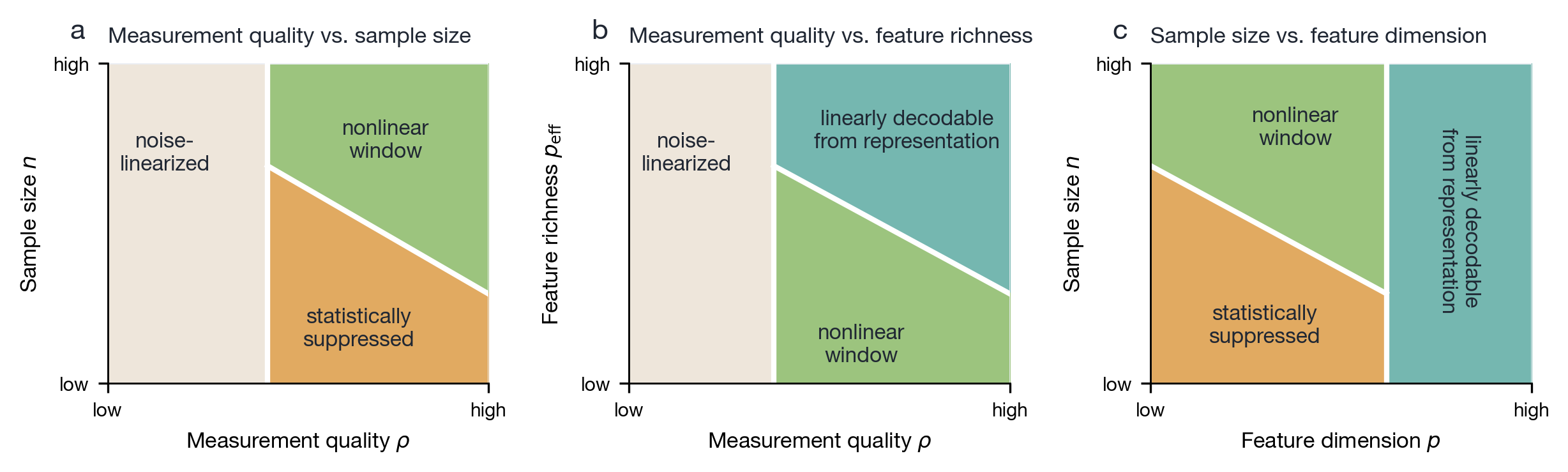}
  \caption{\textbf{The window in which flexible models help.} Three slices through the space of measurement reliability, sample size, and feature dimension. A nonlinear advantage exists only where reliability, sample size, and representation all permit it (green). Its three failure regions are low reliability (noise-linearized), too small a sample (statistically suppressed), and an already-linear representation (linearly decodable). The reliability axis is the latent prediction-relevant reliability; the reported intraclass correlation is an upper bound on it.}
  \label{fig:regime}
\end{figure}

The three look the same in a benchmark, each a flexible model that ties or loses to a linear one, but they are not the same problem. When noise has hidden the nonlinearity, better measurement brings it back, and when the sample is too small, more subjects do; in both, a flexible model can then pull ahead. When the representation has already absorbed the nonlinearity, neither helps, and neither needs to, because the linear model is already optimal on the features at hand. A flexible model gains there only from a different feature set, and only if the old one had thrown signal away. Together the three leave only a narrow window in which a flexible model helps. It is narrower than sample size alone would suggest, because reliability and the representation bound it as tightly and are rarely considered. Most biomedical tasks fall outside it (Fig.~\ref{fig:regime}). 

\section{Evidence from UK Biobank}
\label{sec:evidence}

Equation~\ref{eq:identity} is a statement about population quantities, the error of the best linear and the best possible predictor at infinite sample size. A practitioner instead observes the gap between two fitted models at finite sample size, which is related but not identical. We therefore examine the mechanism first on controlled data, where the population quantities are known, and then on biomedical data, where only the fitted gap is available.

\begin{figure*}[t]
  \centering
  \includegraphics[width=0.92\textwidth]{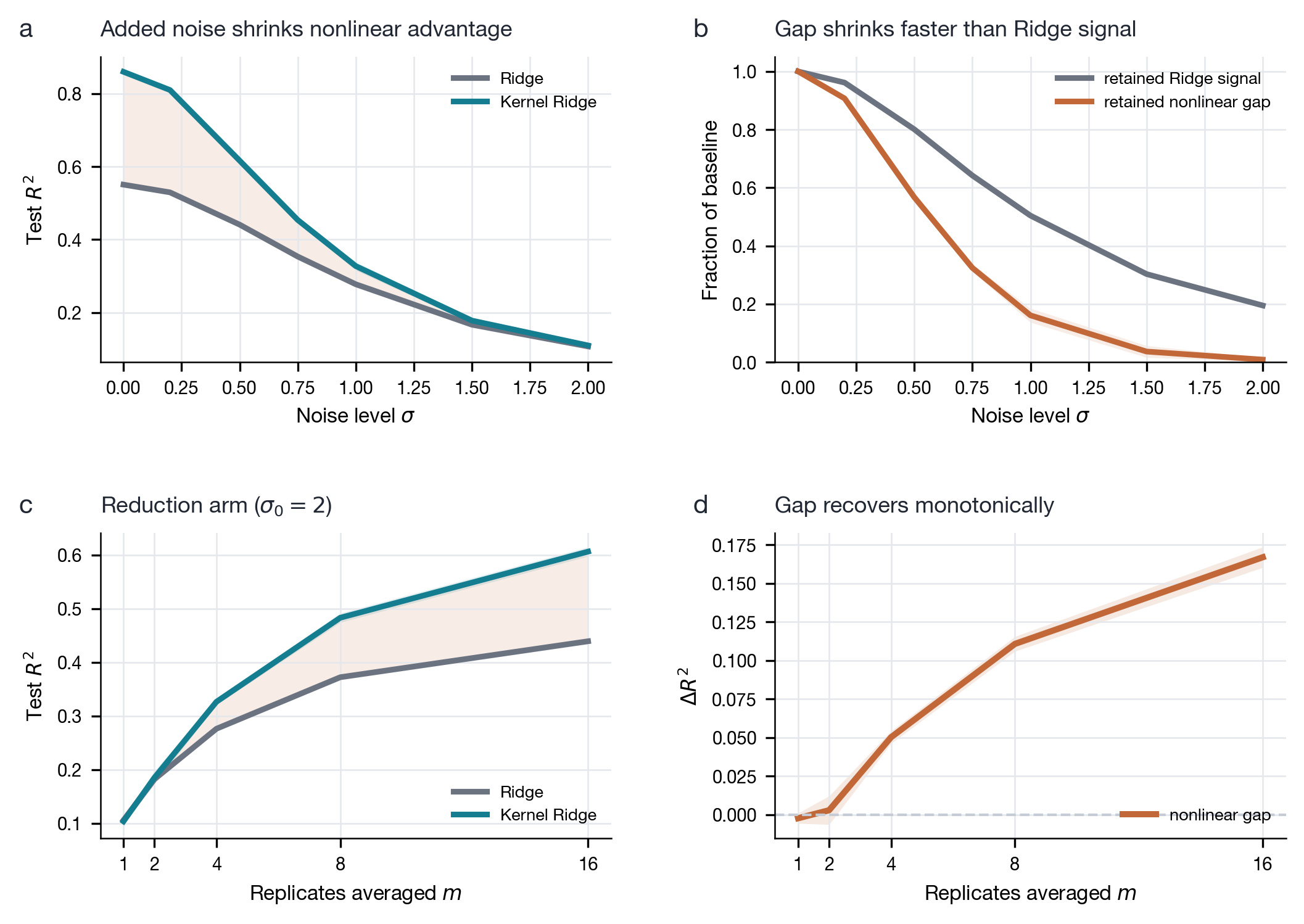}
  \caption{\textbf{Noise hides nonlinear structure, and reducing it brings the structure back.} Ridge regression (grey) and polynomial-kernel ridge (teal) on a controlled benchmark; bands are $\pm 1$ s.e.m.\ across 12~runs. (a)~Both $R^2$ curves fall as noise is added. (b)~The nonlinear gap (orange) falls faster than the linear signal (grey), each normalized to its baseline. (c)~Averaging $m$ replicates restores both curves, with kernel ridge climbing faster. (d)~The gap grows from near zero at $m=1$ as replicates accumulate.}
  \label{fig:synthetic}
\end{figure*}

On a one-dimensional benchmark with a constructed nonlinear target, the two behaviors that Equation~\ref{eq:identity} predicts both appear (Fig.~\ref{fig:synthetic}). Injecting noise into the feature shrinks the gap between kernel ridge and ordinary ridge faster than it shrinks the linear signal, the direction the identity predicts. Averaging repeated noisy measurements, which reduces the noise rather than adding it, brings the gap back. This recovery shows that the nonlinear structure was hidden by the noise rather than absent from the problem, because lowering the noise makes it reappear. In this controlled benchmark the same averaging also rules out the most obvious alternative reading, that noise drives both models down to a shared floor, since a floor would not lift selectively when the noise is averaged away. The surviving gap also sits where the identity says it should. With the features split into a reliable and an unreliable tier, the interactions among the reliable features survive while those touching an unreliable feature decay fastest (Fig.~\ref{fig:selective}), the product-of-reliabilities pattern the identity predicts. The pattern is not specific to kernel ridge; it holds for gradient-boosted trees and multilayer perceptrons (Fig.~\ref{fig:model-class-audit}), on a digit-classification task (Fig.~\ref{fig:mnist}), and under non-Gaussian noise (Fig.~\ref{fig:robustness}); the flexible model is re-tuned at every noise level, so the shrinking gap reflects a simpler optimal predictor rather than a tuning failure (Fig.~\ref{fig:tuning}).

\begin{figure*}[t]
  \centering
  \includegraphics[width=0.92\textwidth]{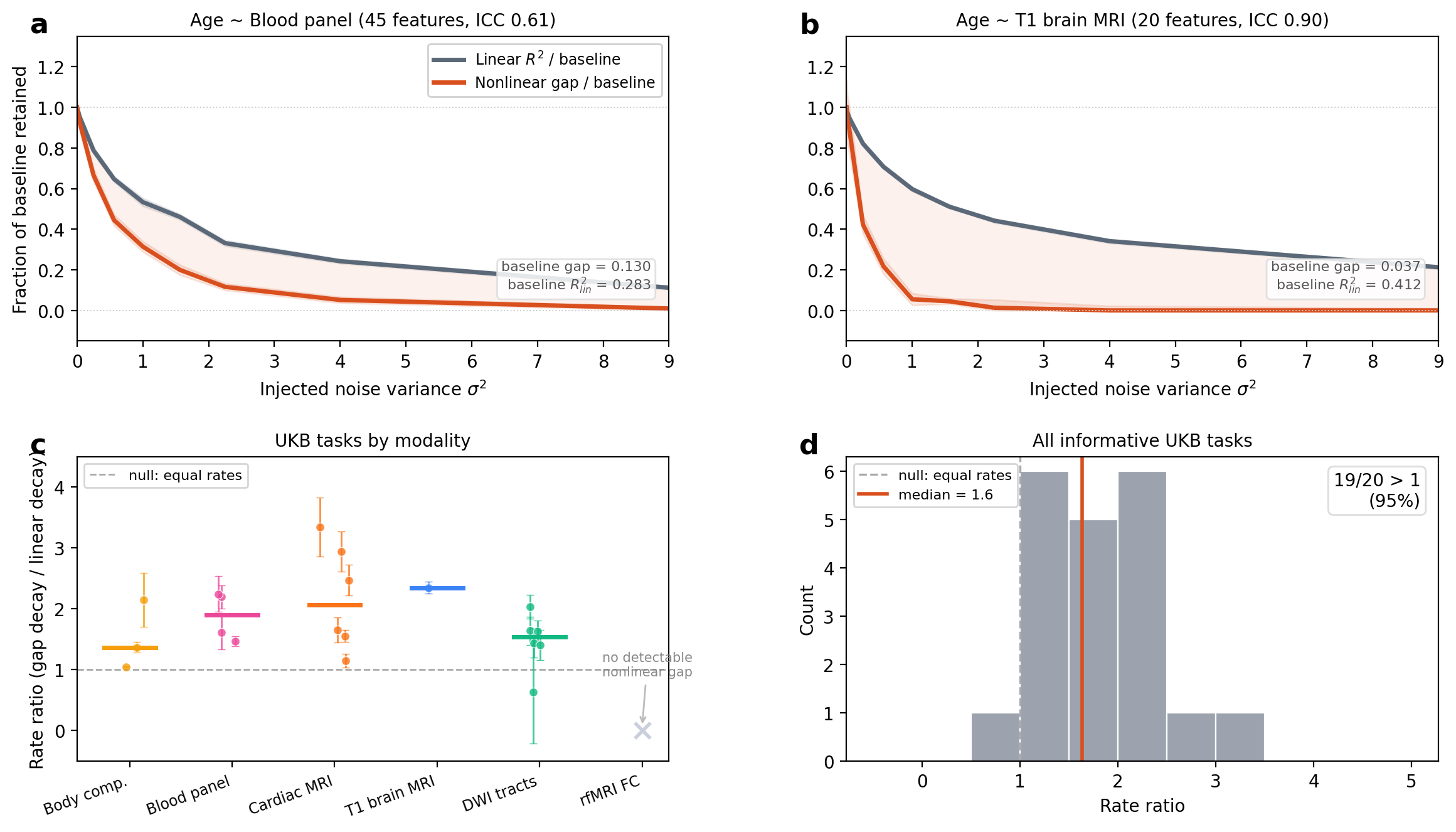}
  \caption{\textbf{The signature on UK Biobank tasks.} (a,b)~Normalized decay curves for age predicted from blood features (45~features, ICC~0.61) and from T1 brain MRI features (20~features, ICC~0.90); in both, the nonlinear gap falls faster than the linear signal. (c)~Decay ratios by modality; resting-state functional connectivity (ICC~0.2--0.3) shows no measurable baseline gap. (d)~Distribution of decay ratios across the 20 informative tasks; 19 lie above one.}
  \label{fig:ukb}
\end{figure*}

The benchmark adds noise to clean features; biomedical features are noisy to begin with. We built 140 prediction tasks from the UK Biobank, each fit at ten thousand participants. The features come from six modalities, body composition, blood biomarkers, cardiac MRI, T1 structural MRI, diffusion MRI, and resting-state functional connectivity; the targets range from continuous phenotypes such as age, body-mass index, and blood pressure to binary disease endpoints such as hypertension and depression. Only 20 of the 140 tasks showed a measurable gap in prediction performance between linear and nonlinear models, consistent with a narrow window in which flexible models help. In the remaining tasks, the two model classes tie even though the linear model predicts well, so the signal is real and linear structure already captures it; only in a minority (30) neither model beats the trivial baseline, so there is no learnable signal for either to find. Resting-state functional connectivity, the modality with the lowest reported reliability, around 0.2 to 0.3 at the level of individual edges \citep{noble2019decade, zuo2014testretest}, produces no nonlinear gap in any of its tasks.
The gaps that do exist behave as the theory predicts. Across the 20 informative tasks, injecting noise shrinks the gap faster than it shrinks the linear signal in 19 (Fig.~\ref{fig:ukb}), the nonlinear advantage proving the more fragile of the two, as differential attenuation requires. The gap between two fitted models is a finite-sample quantity rather than the population excess risk the identity describes, so we read its decay as direction and not rate (\S\ref{sec:si-proxy}).

Comparing the gap across tasks might seem a more direct test, but it is the wrong one. The natural expectation is that the noisiest modalities should show the smallest nonlinear gaps; Equation~\ref{eq:identity} predicts differently. Reliability enters the identity as a multiplier on the clean function's nonlinear structure: it scales that structure but does not supply it. Across tasks as varied as ours, the structure itself ranges from almost none to a great deal, a far wider range than feature reliability spans, so it dominates any comparison across tasks. The gap should therefore track each target's nonlinear predictability rather than its feature reliability. The biobank data follow this prediction: the baseline gap is essentially uncorrelated with modality reliability (Spearman $-0.02$) and instead tracks the strength of the linear signal (Spearman $0.55$). Reliability governs the gap only once the signal is held fixed, which is what the within-task injection sweep does and a cross-task comparison cannot.

An observed small gap therefore reveals nothing about its own cause. A task linearized by noise looks, from the outside, the same as one whose features were never nonlinearly informative and one whose sample was too small to exploit the structure that is present. Inspection cannot separate these cases; intervention can. Adding noise, adding data, or enriching the representation, then seeing which one moves the gap, is what tells them apart, and the injection sweep is the intervention that isolates reliability.

\section{Implications}
\label{sec:record}

The same ambiguity appears at the scale of whole domains. A tie between the two model classes can mean genuinely linear biology, a nonlinearity hidden by feature noise, one left unestimable by too small a sample, or one already absorbed into the features (Table~\ref{tab:domains}). Inspection cannot say which holds, but what is already known about a domain can rule some out. Where features are near-noiseless, measurement is out: a called genotype is almost perfectly reproducible \citep{laurie2010quality}, so the parity of linear and nonlinear genomic prediction is not a measurement effect \citep{kelemen2025performance}. Sample size and representation remain in play instead, since interactions among millions of variants stay unestimable even at the largest samples \citep{hill2008data}, the MHC aside \citep{lenz2015widespread}, and the additive polygenic score has already projected the problem onto its linear part. Continuous-assay omics are not exempt, because the biological test-retest reliability of gene expression and methylation runs far below their technical reliability \citep{shvetsov2015intraindividual}. Where a feature instead folds a nonlinear relationship into a single engineered term, as in imaging-derived phenotypes or estimated GFR, representation is the cause, though such features also discard information. And where features are genuinely unreliable, measurement returns: the edges of functional connectivity, the least reliable features we consider, are consistent with a noise limit and a sample limit at once. The reading narrows the cause in each domain but does not fix it; only intervention can.

\begin{table}[!ht]
\centering
\caption{\textbf{Linear versus nonlinear performance across biomedical domains.} \textsuperscript{\dag}Technical reliability; the prediction-relevant reliability may be lower (\S\ref{sec:implications}). \textsuperscript{\ddag}Engineered features, reproducible by construction. ICC ranges are representative published test-retest reliabilities (sMRI, \citealp{duff2022reliability}; fMRI FC, \citealp{noble2019decade, zuo2014testretest}; Olink, \citealp{haslam2022stability}; methylation and expression, \citealp{pidsley2016critical, maqc2006microarray}).}
\label{tab:domains}
\scriptsize
\setlength{\tabcolsep}{2.5pt}
\renewcommand{\arraystretch}{1.15}
\begin{tabular}{@{}llccccll@{}}
\toprule
Domain & Modality & ICC & $n$ & $d$ & Lin$\approx$NL? & Exception & Source \\
\midrule
Brain img. & sMRI & .66--.93 & 9--50k & 20--1.4k & Yes & --- & \citet{schulz2020different} \\
Brain img. & fMRI FC & .20--.30 & 1--40k & 50--55k & Yes & --- & \citet{he2020deep, han2025rethinking} \\
Proteomics & Olink & .40--.90 & 53k & 2.9k & Mostly & NN 11/27 & \citet{smith2025proteomic} \\
Epigenomics & Methylation & .95--.99\textsuperscript{\dag} & .6--8k & 71--450k & Yes & --- & \citet{horvath2013dna} \\
Genomics & Expression & .95--.99\textsuperscript{\dag} & 50--11k & 2--24k & Yes & --- & \citet{dudoit2002comparison, statnikov2008comprehensive} \\
Clinical & EHR & ${\sim}1.0$\textsuperscript{\ddag} & 1--100k & 10--100 & Yes & --- & \citet{christodoulou2019systematic} \\
Genomics & PGS & ${\sim}1.0$\textsuperscript{\dag} & 10--500k & .1--1M & Yes & --- & \citet{kelemen2025performance} \\
\bottomrule
\end{tabular}
\end{table}

The three conditions that explain the long run of negative results also mark out where a flexible model should win, and in one domain it does. In a large proteomic study, neural networks beat elastic net for 11 of 27 outcomes, with the largest gains for multiple sclerosis, Parkinson's disease, and pulmonary embolism \citep{smith2025proteomic}. Here two of the three conditions are satisfied for reasons that do not depend on the benchmark itself: the assay is reliable, and the sample, at roughly fifty thousand, is large. With reliability and sample size already in hand, the network's advantage is consistent with the third condition, genuine nonlinear structure in the protein-to-disease relationships, making proteomics one of the few documented cases where all three appear to align. Part of that advantage may reflect elastic net's sparsity assumption rather than genuine nonlinearity, which a denser baseline such as ridge would distinguish.

Where instead the measurement is the binding limit, as it often is in biomedicine, the usual responses do not help. More subjects cannot lift a ceiling the noise has already fixed (\S\ref{sec:mechanism}), and more capacity fares no better: a foundation model trained on the same extracted features inherits their blur, since the structure was lost before the model ever saw it. The same smoothing reframes the wider debate over why deep learning trails on tabular data, usually attributed to irregular target functions and network inductive biases \citep{grinsztajn2022why, shwartzziv2021tabular, mcelfresh2023when, borisov2022deep}: the targets look smooth to a network because noise has already smoothed them, before any model is fit. Getting more subjects can even lower the ceiling, because pooling cohorts across sites adds scanner and batch differences that act as further noise, which only harmonization, not sample size, removes.

Which of these causes binds is, for the practitioner, settled only by intervention, through a procedure run once the tie has appeared because the deciding reliability is not visible beforehand. When a small gap is already present, the three interventions of the previous section apply directly, and whichever of added noise, added data, or a richer representation moves the gap names the binding limit. When the two models sit exactly level there is nothing to perturb, and the only way to expose a hidden gap is to raise reliability. Repeating a measurement and averaging lowers the random part of the noise \citep{ooi2025longer}, but not variance that is reproducible yet irrelevant to the target. When a feature is dominated by such variance, stable anatomy or physiology with no bearing on the outcome, averaging cannot help, and the remedy is a different measurement closer to the target.

Several boundaries limit the claim. The evidence comes from complex-trait prediction in one biobank and from published benchmarks of similar tasks, and it shows the mechanism's signature but does not prove that the noise already present is what ties any particular dataset. Injecting noise shows that it erases the nonlinear advantage in the predicted direction; showing that the existing noise is the cause would need that noise lowered, through same-session repeat measurements the UK Biobank does not provide. The identity also describes feature noise alone. Noise in the outcome behaves differently, raising the error of every model together without shrinking the gap between them, so a task limited by noisy labels will not show the signature even where one limited by noisy features would. The additive Gaussian noise of the identity is an idealization as well; real measurement noise, resting-state connectivity again the clearest case, is heterogeneous and partly signal-dependent, departing from the additive-Gaussian model the identity assumes. The mechanism is feature-side and tabular, so it should carry to settings where features are tabular and their noise is roughly additive, and not obviously to raw sensor and time-series data, where the representation question comes first.

When a flexible model fails to beat a linear one, the common response is to treat the model or the sample as the problem and to reach for a stronger model, more subjects, or finer tuning. The limit can instead lie in the measurement. Where features have lost their nonlinear structure to noise, no model and no larger cohort recovers it, and only a better measurement can. The same account points to where the effort is most likely to pay, where reliable features, a large sample, and genuine nonlinear structure meet, as they plausibly do in the proteomic case. Because a benchmark alone cannot tell these cases apart, reporting the test-retest reliability of features alongside sample size and dimension would let the field weigh a null result for flexible models against the quality of the measurement before reading it as simple biology. The window in which flexible models help is narrow in today's biomedical data, set by reliability as much as by sample size, and better measurement widens it where more subjects cannot.

\section*{Acknowledgements}
We thank Lydia Federmann, Sam Gijsen, Didem Stark, and Moritz Seiler for feedback on the draft. Funded by the Deutsche Forschungsgemeinschaft (DFG, German Research Foundation) -- 414984028 to K.R.\ and 2797/9-1 to M.A.S. We also thank the Gemeinn\"{u}tzige Hertie-Stiftung. This research has been conducted using the UK Biobank Resource under Application Number 33073.

During the preparation of this work the authors used a large language model (Anthropic's Claude) to assist with editing the manuscript and with analysis code. The authors reviewed and verified all content and take full responsibility for the work.

\bibliographystyle{plainnat}
\bibliography{references_perspective}

\clearpage
\appendix
\renewcommand{\thesection}{S\arabic{section}}
\renewcommand{\thefigure}{S\arabic{figure}}
\renewcommand{\thetable}{S\arabic{table}}
\renewcommand{\theHsection}{S\arabic{section}}
\renewcommand{\theHfigure}{S\arabic{figure}}
\renewcommand{\theHtable}{S\arabic{table}}
\setcounter{section}{0}
\setcounter{figure}{0}
\setcounter{table}{0}

\begin{center}
{\Large\bfseries Supplemental Information}\\[0.5em]
{\large Measurement noise limits the advantage of nonlinear models over linear models in biomedical prediction}
\end{center}

\vspace{1em}

\section{Derivation of Identity R1 (Hermite excess-risk identity)}
\label{sec:si-r1}

\subsection{Setup}

The main text states R1 for standardized features $X \sim N(0, I_d)$. Here we give the general version for $X \sim N(\mu, \Lambda)$, which reduces to the main-text form when $\mu = 0$ and $\Lambda = I_d$ (so that $\rho_j = 1/(1 + \sigma_j^2)$).

Let $X \sim N(\mu, \Lambda)$ with $\Lambda = \mathrm{diag}(\lambda_1, \ldots, \lambda_d)$, $\lambda_j > 0$. Let $Y = f(X) + \varepsilon$ where $f \in L^2(N(\mu, \Lambda))$ and $\varepsilon$ is independent label noise. Observations are $\tilde{X} = X + \eta$ with $\eta \sim N(0, \Sigma_\eta)$, $\Sigma_\eta = \mathrm{diag}(\sigma_1^2, \ldots, \sigma_d^2)$, independent of $(X, Y)$.

Define the per-feature reliability $\rho_j := \lambda_j / (\lambda_j + \sigma_j^2) \in (0,1)$.

Expand $f$ in the orthonormal Hermite basis adapted to $N(\mu, \Lambda)$:
\[
f(x) = \sum_{\alpha \in \mathbb{N}_0^d} c_\alpha \, h_\alpha\!\left(\Lambda^{-1/2}(x - \mu)\right)
\]
where $h_\alpha(z) = \prod_{j=1}^d \mathrm{He}_{\alpha_j}(z_j) / \sqrt{\alpha_j!}$ are normalized probabilist Hermite polynomials satisfying $\langle h_\alpha, h_\beta \rangle_{L^2(\gamma)} = \delta_{\alpha\beta}$, and $c_\alpha = E[f(X) \, h_\alpha(\Lambda^{-1/2}(X-\mu))]$.

\subsection{Derivation}

The derivation uses Mehler's formula, the eigenvalue identity of the Ornstein--Uhlenbeck semigroup on Gaussian space.

\textbf{Step 1. Standardize.} Define $U = \Lambda^{-1/2}(X - \mu) \sim N(0, I_d)$ and $V = (\Lambda + \Sigma_\eta)^{-1/2}(\tilde{X} - \mu) \sim N(0, I_d)$.

\textbf{Step 2. Bivariate structure.} The pair $(U_j, V_j)$ is jointly Gaussian with $\mathrm{Corr}(U_j, V_j) = \sqrt{\rho_j}$. We can therefore decompose $U_j = \sqrt{\rho_j} \, V_j + \sqrt{1 - \rho_j} \, W_j$ where $W_j \perp V_j$ is standard normal.

\textbf{Step 3. Mehler's formula.} The Ornstein--Uhlenbeck noise operator satisfies:
\[
E[\mathrm{He}_n(U_j) \mid V_j] = \rho_j^{n/2} \, \mathrm{He}_n(V_j)
\]
This is the eigenvalue identity: each Hermite polynomial of degree $n$ is an eigenfunction of the conditional expectation operator with eigenvalue $\rho_j^{n/2}$.

\textbf{Step 4. Product structure.} Because $\Lambda$ and $\Sigma_\eta$ are both diagonal, the coordinates are conditionally independent, so:
\[
E[h_\alpha(U) \mid V] = \prod_{j=1}^d \rho_j^{\alpha_j / 2} \, h_\alpha(V)
\]

\textbf{Step 5. Assembly.} By linearity:
\[
g^*(\tilde{x}) := E[Y \mid \tilde{X} = \tilde{x}] = \sum_\alpha c_\alpha \prod_j \rho_j^{\alpha_j / 2} \, h_\alpha\!\left((\Lambda + \Sigma_\eta)^{-1/2}(\tilde{x} - \mu)\right)
\]
This is R1 part (a): each degree-$|\alpha|$ Hermite component of the Bayes-optimal predictor is attenuated by $\prod_j \rho_j^{\alpha_j/2}$.

The best linear predictor of $Y$ given $\tilde{X}$ uses only the $|\alpha| \le 1$ terms of $g^*$, giving R1 part (b).

By Parseval's identity (orthonormality of the Hermite basis under the Gaussian measure on $V$):
\[
R_{\mathrm{lin}}^* - R_{\mathrm{Bayes}} = \sum_{|\alpha| \ge 2} c_\alpha^2 \prod_{j=1}^d \rho_j^{\alpha_j}
\]
This is R1 part (c). Each nonlinear mode contributes $c_\alpha^2 \prod_j \rho_j^{\alpha_j}$ to the excess risk, where the $\rho_j^{\alpha_j}$ factor arises from squaring the per-coefficient attenuation $\rho_j^{\alpha_j/2}$. \hfill$\square$

\textbf{Note on the exponent convention.} The excess risk formula uses $\rho_j^{\alpha_j}$ (not $\rho_j^{2\alpha_j}$) because the $c_\alpha$ are the Hermite coefficients of the \emph{clean} regression function $f$, not of the noisy Bayes predictor $g^*$. The attenuation $\rho_j^{\alpha_j/2}$ per coefficient is squared in the energy sum, yielding $\rho_j^{\alpha_j}$.

\subsection{Coefficient versus energy attenuation}

Two related attenuation factors should not be confused:

\begin{center}
\small
\begin{tabular}{@{}lcc@{}}
\toprule
Object & Per mode $\alpha$ & Isotropic \\
\midrule
Hermite coefficient of $g^*$ & $\prod_j \rho_j^{\alpha_j/2}$ & $\rho^{|\alpha|/2}$ \\
Excess risk contribution (energy) & $\prod_j \rho_j^{\alpha_j}$ & $\rho^{|\alpha|}$ \\
\bottomrule
\end{tabular}
\end{center}

\noindent The excess risk formula uses the energy attenuation $\prod_j \rho_j^{\alpha_j}$. At $\rho = 0.5$, a pairwise interaction ($|\alpha| = 2$) retains 25\% of its energy; a three-way interaction retains 12.5\%.

\subsection{Conditions}

\begin{center}
\small
\setlength{\tabcolsep}{3pt}
\begin{tabular}{@{}lcll@{}}
\toprule
Condition & Required? & Role & Relaxation \\
\midrule
$f \in L^2(N(\mu, \Lambda))$ & Yes & Hermite convergence & Minimal \\
$X$ Gaussian & Yes (exact) & Mehler's formula & Qualitative for general $X$ \\
$\Lambda, \Sigma_\eta$ diagonal & Yes & Coordinate-wise Mehler & Isotropic $\sigma^2 I$ commutes with any $\Lambda$ \\
$\eta$ Gaussian & Yes (exact) & Convolution closure & Qualitative for finite-variance $\eta$ \\
$\eta \perp (X,Y)$ & Yes & Additive noise model & Breaks for signal-dependent noise \\
\bottomrule
\end{tabular}
\end{center}

A word on the diagonality condition, since it bears on the selective-linearization reading in the main text. The clean per-feature reliability product $\prod_j \rho_j^{\alpha_j}$ requires that $\Lambda$ and $\Sigma_\eta$ be simultaneously diagonalizable in the feature basis, which holds when features are uncorrelated, or when the injected noise is isotropic ($\Sigma_\eta = \sigma^2 I$, commuting with any $\Lambda$). When features are correlated \emph{and} noise is feature-specific, the identity still holds but in the eigenbasis of the joint structure, not the raw feature basis, so ``per-feature reliability'' becomes an approximation. The empirical noise-injection experiments use isotropic injected noise and so fall in the exact case.

\subsection{Rates}

The leading nonlinear degree $k_{\min}(f) := \min\{k \ge 2 : \exists |\alpha| = k, \, c_\alpha \ne 0\}$ determines the asymptotic rate. Under isotropic noise with $\sigma^2 \gg \max_j \lambda_j$: the excess risk decays as $O(1/\sigma^{2k_{\min}})$. For most functions of practical interest, $k_{\min} = 2$, giving $O(1/\sigma^4)$. Odd functions (e.g., $\sin$, $\tanh$, step functions) have $k_{\min} = 3$ and decay as $O(1/\sigma^6)$.

\subsection{Connection to ICC}

Under the additive noise model $\tilde{X} = X + \eta$ with $X \perp \eta$:
\[
\rho_j = \frac{\mathrm{Var}(X_j)}{\mathrm{Var}(\tilde{X}_j)} = \mathrm{ICC}_j
\]
This is definitional: $\mathrm{ICC}_j$ is the ratio of between-subject variance to total variance. The excess risk in ICC terms:
\[
R_{\mathrm{lin}}^* - R_{\mathrm{Bayes}} = \sum_{|\alpha| \ge 2} c_\alpha^2 \prod_{j=1}^d \mathrm{ICC}_j^{\alpha_j}
\]
As discussed in \S\ref{sec:implications} of the main text, the prediction-relevant reliability $\rho_{\mathrm{rel},j}$ may be substantially smaller than $\mathrm{ICC}_j$ when reproducible but target-irrelevant variance is present.

\section{Bound C1 (Classification coupling bound)}
\label{sec:si-c1}

Unlike R1, which is an exact identity, the classification result is a one-sided upper bound. It is enough to establish that the optimal decision boundary linearizes as noise grows, driving the excess risk of the best linear discriminant to zero, but it does not pin the rate down as tightly, and its dominant term decays only as $1/\sigma$. We state it, prove the bound, and then comment on its tightness.

\subsection{Setting}

$K \ge 2$ classes with priors $\pi_k > 0$. Each class $k$ has a Gaussian mixture class-conditional distribution:
\[
p(x \mid k) = \sum_{j=1}^{J_k} w_{kj} \, N(\mu_{kj}, \Sigma_{kj})
\]
with diagonal component covariances $\Sigma_{kj} = \mathrm{diag}(\sigma^2_{kj,1}, \ldots, \sigma^2_{kj,d})$. Observations: $\tilde{X} = X + \eta$, $\eta \sim N(0, \Sigma_\eta)$, $\Sigma_\eta = \mathrm{diag}(\sigma_{\eta,1}^2, \ldots, \sigma_{\eta,d}^2)$, independent of $X$.

\subsection{Statement}

Define the pooled noisy covariance $B := \mathrm{diag}(\bar{\Sigma}_{ii} + \sigma_{\eta,i}^2)$ where $\bar{\Sigma} = \sum_k \sum_j \pi_k w_{kj} \Sigma_{kj}$. Under the per-feature noise dominance condition ($\sigma_{\eta,i}^2 \ge 2 \max_{k,j} |\Sigma_{kj,ii} - \bar{\Sigma}_{ii}|$ for all $i$):
\[
R_{\mathrm{lin}}^*(\Sigma_\eta) - R_{\mathrm{Bayes}}(\Sigma_\eta) \le \sum_k \pi_k \left(\|\delta_k\| + \|M_k\|_F\right)
\]
where $\|\delta_k\|^2 := \sum_j w_{kj} \sum_i (\mu_{kj,i} - \bar{\mu}_{k,i})^2 / (\bar{\Sigma}_{ii} + \sigma_{\eta,i}^2)$ captures within-class mean scatter, and $\|M_k\|_F^2 := \sum_j w_{kj} \sum_i [(\Sigma_{kj,ii} - \bar{\Sigma}_{ii}) / (\bar{\Sigma}_{ii} + \sigma_{\eta,i}^2)]^2$ captures covariance heterogeneity.

\subsection{Two mechanisms, two rates}

The bound has two interpretable terms with different decay rates. The $\|\delta_k\|$ term (component merging) captures within-class mixture components becoming indistinguishable under noise and decays as $O(1/\sigma)$. The $\|M_k\|_F$ term (covariance equalization) captures cross-class covariance differences being washed out and decays as $O(1/\sigma^2)$. When all classes share the same covariance and each is a single Gaussian, both terms vanish and Fisher's LDA is exactly Bayes-optimal.

\textbf{On tightness and the asymmetry with R1.} Two points qualify the comparison with R1. First, because the bound is an upper bound, the true excess risk may decay faster than these rates suggest; the $1/\sigma$ and $1/\sigma^2$ figures are guarantees, not measured rates. Second, the slower ($1/\sigma$) term dominates, so the boundary-curvature intuition emphasized in Fig.~\ref{fig:framework}b, which corresponds to the faster covariance-equalization term, is not what controls the bound's leading behavior; within-class subtype heterogeneity is. The regression case (R1) is therefore both exact and faster ($1/\sigma^4$), whereas the classification case is a one-sided guarantee with a slower leading term. We present C1 as establishing the qualitative linearization for classification, with R1 carrying the quantitative weight of the paper.

\subsection{Proof sketch}

\begin{enumerate}
\item Construct a reference model $q_k := N(\bar{\mu}_k, B)$ with shared covariance. Fisher's LDA is Bayes-optimal for $\{q_k, \pi_k\}$.
\item Coupling inequality: $R_{\mathrm{lin}}^* - R_{\mathrm{Bayes}} \le 2 \sum_k \pi_k \, \mathrm{TV}(p_k, q_k)$.
\item Pinsker's inequality: $\mathrm{TV}(p_k, q_k) \le \sqrt{\mathrm{KL}(p_k \| q_k) / 2}$.
\item KL convexity: $\mathrm{KL}(p_k \| q_k) \le \sum_j w_{kj} \, \mathrm{KL}(p_{kj} \| q_k)$.
\item Diagonal Gaussian KL: under noise dominance, $r_i - \ln(1 + r_i) \le r_i^2$, yielding the Frobenius norm.
\item Apply $\sqrt{a + b} \le \sqrt{a} + \sqrt{b}$ to separate the mean and covariance terms. \hfill$\square$
\end{enumerate}

\section{A $\sigma^4$ sample-complexity heuristic}
\label{sec:si-sigma4}

This section gives the scaling argument behind the main text's statement that the data requirement grows as $D \cdot \sigma^4$. We present it as a heuristic, not a theorem: it combines the \emph{exact} bias term from R1 with a \emph{generic} estimation rate, and, as we make explicit below, the two are quantified over different objects, so the resulting expression is a rule of thumb for how the sample requirement scales with noise, not a sharp lower bound on the sample size for a given problem.

\subsection{The scaling argument}

\textbf{Step 1 (Bias shrinkage from R1, exact).} From Identity R1, the excess risk of linearity at large $\sigma$ is dominated by the degree-2 terms:
\[
\varepsilon_0 := R_{\mathrm{lin}}^* - R_{\mathrm{Bayes}} = \frac{\tilde{C}_2}{\sigma^4} + O(1/\sigma^6),
\qquad
\tilde{C}_2 := \sum_{|\alpha|=2} c_\alpha^2 \prod_j \lambda_j^{\alpha_j}.
\]
This is the exact degree-2 Hermite energy of $f$, weighted by signal variances.

\textbf{Step 2 (Estimation rate, generic).} By standard minimax theory (e.g., Fano's method or metric-entropy arguments; see Tsybakov 2009), estimation over a model class $\mathcal{F}$ of pseudo-dimension $D$ obeys
\[
\inf_{\hat{f}_n} \sup_{g \in \mathcal{F}} E\!\left[\|\hat{f}_n - g\|^2\right] \ge \frac{c \cdot D}{n}.
\]

\textbf{Step 3 (Combining, heuristic).} Setting the estimation scale equal to the bias gap, a nonlinear estimator can plausibly improve on the best linear predictor once $c D / n \lesssim \varepsilon_0 = \tilde{C}_2/\sigma^4$, i.e. once
\[
n \gtrsim n^* \sim \frac{c \cdot D \cdot \sigma^4}{\tilde{C}_2}.
\]

\textbf{Consequence (the part we rely on).} Whatever the constants, the $\sigma^4$ dependence means that doubling the measurement noise ($\sigma \to 2\sigma$) inflates the sample requirement by a factor of about 16. It is this \emph{scaling}, not the absolute constant, that the main text invokes.

\subsection{Where the argument falls short of a theorem}

Two gaps separate the argument above from a rigorous lower bound on $n^*$.

\emph{(i) Mismatched quantifiers.} The estimation bound in Step 2 is a \emph{minimax} statement: it says there exists a hard $g \in \mathcal{F}$ requiring $n \gtrsim cD/\varepsilon$ samples. It does not say that the \emph{specific} target $g^*$ whose energy is $\tilde{C}_2$ is that hard; $g^*$ may be far easier to estimate. Combining a worst-case-over-class estimation rate with a specific-target bias term mixes quantifiers, so the resulting $n^*$ is a scaling guide rather than a guarantee for a given problem.

\emph{(ii) The linear baseline also has variance.} The argument compares the nonlinear estimator's error against the bias gap but ignores that the \emph{linear} estimator is itself fit from data and carries its own (smaller) estimation variance. A complete accounting would compare nonlinear bias-plus-variance against linear bias-plus-variance; the omission is in the conservative direction but is an omission nonetheless.

For these reasons the $\sigma^4$ scaling is the one quantitative claim of the paper that we do \emph{not} formalize in Lean. It is included because the $\sigma^4$ dependence is robust to the constants and is practically useful, not because the derivation is airtight.

\subsection{Compounding effect of noise on sample requirements}

Read alongside R1, the scaling captures a two-sided squeeze on the nonlinear window. First, R1 shows that noise shrinks the bias gap $\varepsilon_0$ that a nonlinear model could exploit. Second, the $\sigma^4$ scaling means the sample size required to exploit whatever gap remains grows rapidly with noise. Fig.~S6 illustrates this compounding on the synthetic benchmark: the payoff from increasing sample size peaks at intermediate noise levels and declines at high noise, where the problem has been linearized and additional data cannot recover the nonlinear signal.

\section{UK Biobank noise injection experiment}
\label{sec:si-ukb}

\subsection{Task definitions}

We constructed 140 prediction task--modality combinations from the UK Biobank, crossing six feature modalities with 12 regression targets and 11--12 classification targets. Table~\ref{tab:modalities} lists the feature modalities and Table~\ref{tab:targets} lists all prediction targets.

\begin{table}[htbp]
\centering
\scriptsize
\setlength{\tabcolsep}{3pt}
\caption{Feature modalities.}
\label{tab:modalities}
\begin{tabular}{llrlll}
\toprule
Modality & Description & $d$ & Published ICC & UKB field range \\
\midrule
Body composition & DXA-derived impedance measures & 31 & 0.80--0.95 & fid-23099 to fid-23130 (t0) \\
Blood biomarkers & Blood biochemistry + haematology panel & 45 & 0.50--0.85 & fid-30000 to fid-30890 (t0) \\
Cardiac MRI & LV, RV, LA, RA structure and function & 21 & 0.60--0.85 & fid-22420 to fid-22426 (t2) \\
T1 structural MRI & Global volumes + bilateral subcortical & 20 & 0.66--0.90 & fid-25000 to fid-25024 (t2) \\
Diffusion MRI & FA + MD tract-skeleton measures & 96 & 0.60--0.85 & fid-25056 to fid-25152 (t2) \\
Resting-state fMRI & ICA-100 partial corr.\ (50 PCs) & 50 & 0.20--0.30 & rfmri-pcorr (t2) \\
\bottomrule
\end{tabular}

\vspace{0.5em}
\raggedright
Each modality represents a curated subset of available UKB fields selected for measurement quality and biological coverage, not the full set of available features. Feature sets were fixed before any analysis. The resting-state functional-connectivity features are the first 50 principal components of the 1,485 ICA-100 partial correlations, computed on the training split before any noise injection to keep the RBF kernel tractable at this dimensionality. The published reliability range (0.20 to 0.30) refers to the raw partial-correlation edges; the reliability of the principal components was not separately estimated.
\end{table}

\begin{table}[htbp]
\centering
\small
\caption{Prediction targets.}
\label{tab:targets}
\begin{tabular}{llll}
\toprule
Category & Target & Type & UKB field / ICD code \\
\midrule
General & Age & reg & fid-21003 \\
General & Reaction time & reg & fid-20023 \\
General & Sex & clf & fid-31 \\
General & Smoking (ever/never) & clf & fid-20116 \\
Metabolic & BMI & reg & fid-21001 \\
Metabolic & HbA1c & reg & fid-30750 \\
Metabolic & Hypothyroidism & clf & icd-E03 \\
Metabolic & Dyslipidemia & clf & icd-E78 \\
Cardiovascular & Systolic blood pressure & reg & fid-4080 \\
Cardiovascular & Pulse rate & reg & fid-102 \\
Cardiovascular & Hypertension & clf & icd-I10 \\
Cardiovascular & Angina pectoris & clf & icd-I20 \\
Musculoskeletal & Grip strength & reg & fid-46 \\
Musculoskeletal & Heel bone density & reg & fid-3148 \\
Musculoskeletal & Hip osteoarthritis & clf & icd-M16 \\
Musculoskeletal & Osteoporosis & clf & icd-M81 \\
Respiratory & Forced vital capacity & reg & fid-3063 \\
Respiratory & FEV1 & reg & fid-3062 \\
Respiratory & COPD & clf & icd-J44 \\
Respiratory & Asthma & clf & icd-J45 \\
Psychiatric & Neuroticism score & reg & fid-20127 \\
Psychiatric & Sleep duration & reg & fid-1160 \\
Psychiatric & Depression & clf & icd-F32 \\
Psychiatric & Anxiety & clf & icd-F41 \\
\bottomrule
\end{tabular}
\end{table}

Disease endpoints were defined as prevalent ICD-10 diagnoses from hospital episode statistics. Classification tasks used case-control matching with a 3:1 control-to-case ratio.

The task ``HbA1c predicted from blood biomarkers'' was excluded from the rate-mismatch analysis because HbA1c is itself a blood biomarker, so features and target share measurement error, violating the assumption $\eta \perp Y$. This single exclusion leaves 139 task--modality combinations for the rate-mismatch analysis, of which 20 met the informativeness criteria below.

\subsection{Model specifications}

\textbf{Linear model:} ridge regression (\texttt{sklearn.linear\_model.Ridge}) for regression tasks; ridge classifier (\texttt{sklearn.linear\_model.RidgeClassifier}) for classification tasks.

\textbf{Nonlinear model:} RBF kernel ridge regression (\texttt{sklearn.kernel\_ridge.KernelRidge} with \texttt{kernel="rbf"}) for regression; a custom kernel ridge classifier (one-vs-rest with $\{-1, +1\}$ encoding) for classification.

\subsection{Model generality}

The main-text results use ridge versus RBF kernel ridge as the estimator pair. To verify that the rate-mismatch finding is not specific to this choice, we repeated the synthetic benchmark (Fig.~\ref{fig:synthetic}) with three additional nonlinear estimators: XGBoost (100 trees, max depth 4), a multilayer perceptron (2 hidden layers, 128 units, ReLU), and a second ridge regression (as a negative control). All three nonlinear models show the same pattern: the nonlinear gap decays faster than the linear signal under noise injection, and averaging replicates restores the gap at the mirror rate (Fig.~S1). The ridge-versus-ridge control confirms equal decay rates, as expected. R1 predicts this generality: if $E[Y \mid \tilde{X}]$ is approximately linear, no nonlinear estimator can substantially outperform ridge.

\subsection{Tuning behavior under noise injection}

\begin{figure}[htbp]
\centering
\includegraphics[width=\textwidth]{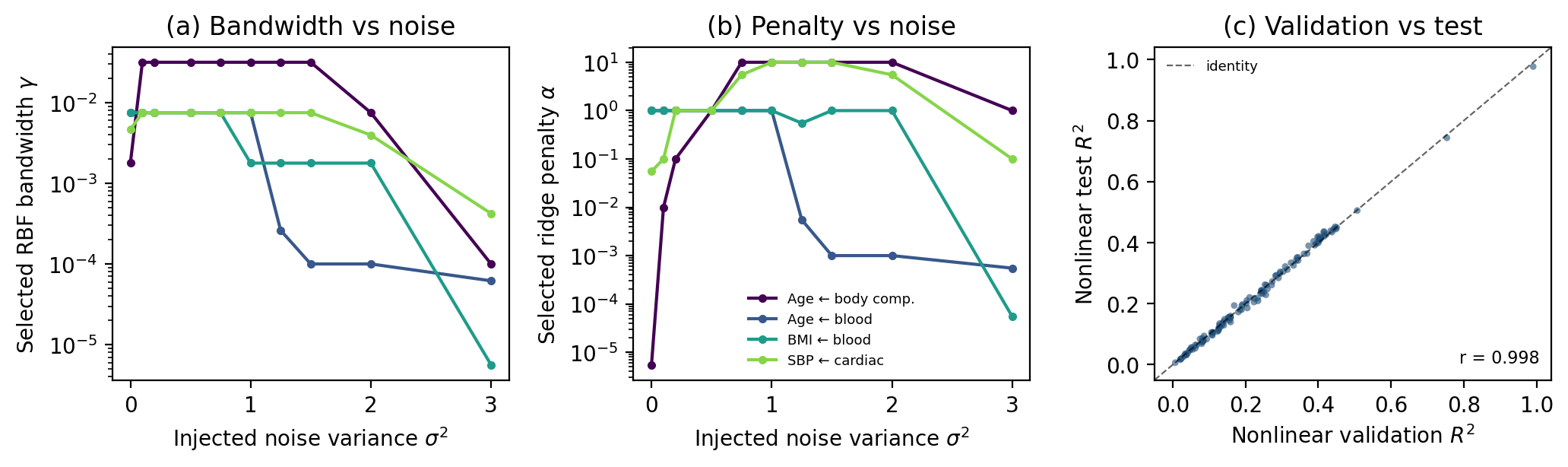}
\caption{\textbf{Hyperparameter selection under injected noise.} The RBF kernel ridge model is re-tuned by validation at every injected-noise level, shown here for four representative tasks. (a)~The selected RBF bandwidth $\gamma$ falls as noise grows, widening the kernel toward a smoother function. (b)~The selected ridge penalty $\alpha$ rises. (c)~Across all informative regression tasks and noise levels, the nonlinear model's validation $R^2$ tracks its test $R^2$ almost exactly (Pearson $r = 0.998$; points lie on the identity line), so the validation-based tuning generalizes and the model is not under-fit at high noise. The gap collapse reflects a simpler optimal predictor that tuning recovers.}
\label{fig:tuning}
\end{figure}

The nonlinear model is re-tuned by validation at every noise level. As noise grows, the selected RBF bandwidth $\gamma$ falls, widening the kernel toward a smoother function, and the ridge penalty $\alpha$ rises (Fig.~\ref{fig:tuning}a,b); the model is tuned toward the simpler function that minimizes validation error. Across all informative tasks and noise levels the nonlinear model's validation $R^2$ tracks its test $R^2$ almost exactly (Pearson $r = 0.998$; Fig.~\ref{fig:tuning}c), so the model is neither over- nor under-fit. The shrinking gap therefore reflects a simpler optimal predictor that tuning recovers correctly.

\subsection{Hyperparameter grids}

Hyperparameters were selected by grid search on a held-out validation set.

\textbf{ridge regularization} $\alpha$: 17 values on a log-uniform grid from $10^{-7}$ to $10^5$ (base grid of 5 points from $10^{-5}$ to $10^3$ with 2 refinement subdivisions and 2 expansion points at each end).

\textbf{Kernel ridge regularization} $\alpha$: same 17-value grid as ridge.

\textbf{RBF bandwidth} $\gamma$: 17 values on a log-uniform grid from $10^{-7}$ to $10^2$ (base grid of 5 points from $10^{-4}$ to $10^1$ with refinement and expansion). The nonlinear model searches over the cross-product of $\alpha$ and $\gamma$ grids (289 combinations per noise level).

\subsection{Noise injection protocol}

For each task, we:
\begin{enumerate}
\item Load features and targets, dropping rows with missing values.
\item Standardize features to zero mean and unit variance on the training set.
\item For each noise level $\sigma^2 \in \{0, 0.1, 0.2, 0.5, 0.75, 1.0, 1.25, 1.5, 2.0, 3.0\}$:
  \begin{enumerate}
  \item Add independent Gaussian noise $\eta_{ij} \sim N(0, \sigma^2)$ to each standardized feature.
  \item Re-standardize the noisy features (to maintain unit variance for the kernel).
  \item Select hyperparameters on the validation set.
  \item Evaluate on the held-out test set.
  \end{enumerate}
\item Repeat steps 1--3 across 10 independent train/validation/test splits.
\end{enumerate}

\textbf{Split sizes:} $n_{\mathrm{train}} = 10{,}000$, $n_{\mathrm{val}} = 1{,}000$, $n_{\mathrm{test}} = 1{,}000$. Standardization parameters (mean, variance) are computed on the training set only and applied to validation and test sets to prevent information leakage.

\textbf{Re-standardization note.} After adding noise with variance $\sigma^2$ to standardized features (variance 1), the total variance is $1 + \sigma^2$. Re-standardizing divides by $\sqrt{1 + \sigma^2}$, giving effective features $\tilde{X}_{\mathrm{std}} = (X + \eta) / \sqrt{1 + \sigma^2}$. This preserves the reliability structure: $\rho_j = \mathrm{Var}(X_j) / \mathrm{Var}(\tilde{X}_j) = 1 / (1 + \sigma^2)$, matching R1's definition. Re-standardization is necessary to keep the RBF kernel bandwidth comparable across noise levels. The injected noise is isotropic across features, placing the experiment in the exact case of R1 (\S\ref{sec:si-r1}).

\textbf{Metrics:} $R^2$ for regression, accuracy for classification.

\subsection{Informativeness threshold}

A task is classified as ``informative'' (having a detectable baseline nonlinear gap) if all four conditions are met at $\sigma^2 = 0$:
\begin{enumerate}
\item \textbf{Margin:} Both linear and nonlinear scores exceed the trivial floor (mean prediction for regression, majority class for classification) by at least 0.02.
\item \textbf{Headroom:} The maximum score across both models exceeds the floor by at least 0.05.
\item \textbf{Relative gap:} $|\text{gap}| / \text{headroom} \ge 0.05$ (gap is at least 5\% of the available signal).
\item \textbf{Significance:} $\text{gap} / \mathrm{SE}_{\text{gap}} \ge 2.0$ (gap exceeds two standard errors across the 10 repeats).
\end{enumerate}

Of the 139 task--modality combinations entering the rate-mismatch analysis (after the same-panel exclusion), 20 met all four criteria, and all 20 are regression tasks; no classification task crossed the threshold. The nonlinear gap decays faster than the linear signal (decay ratio $>1$) in 19 of these 20. The single exception, systolic blood pressure predicted from diffusion MRI, has a baseline gap of $0.007$ in $R^2$: distinguishable from zero ($z = 4.1$) but small enough that its decay ratio is poorly determined.

\textbf{Sensitivity analysis.} Relaxing the significance threshold from $z \ge 2.0$ to $z \ge 1.5$ adds tasks with decay ratios above 1; tightening to $z \ge 2.5$ removes borderline tasks. The direction tally (large majority with decay ratio $>1$) is stable across thresholds.

\subsection{Decay ratio: a direction statistic}

For each informative task we summarize the asymmetry between how the nonlinear gap and the linear signal respond to noise, using the values measured directly at $\sigma^2 = 1$, the point where the injected noise variance equals the standardized feature variance ($\rho = 0.5$). Let $\Delta(\sigma^2) = \text{score}_{\text{nonlinear}} - \text{score}_{\text{linear}}$ be the nonlinear gap and $L(\sigma^2) = \text{score}_{\text{linear}}(\sigma^2)$ the linear signal. The retained fractions at $\sigma^2 = 1$ are $\Delta_{\text{retained}} = \Delta(1)/\Delta(0)$ and $L_{\text{retained}} = L(1)/L(0)$, both read from the measured curves rather than from a parametric fit.

The \textbf{decay ratio} is $(1 - \Delta_{\text{retained}}) / (1 - L_{\text{retained}})$. A value of 1.0 means noise damages the nonlinear gap and the linear signal equally; values above 1 mean the nonlinear gap is more noise-sensitive, the direction R1 predicts. We read the decay ratio as a \emph{direction} statistic (is the nonlinear gap preferentially destroyed?) and not as an estimate of R1's exponent. We use the measured retained fractions rather than a fitted curve precisely because, as the next subsection shows, the fitted exponent is an unreliable proxy for the population rate. Error bars reflect the variation of the retained fractions across the 10 repeats.

\subsection{The fitted exponent is a biased proxy for R1's population rate}
\label{sec:si-proxy}

It is tempting to read the fitted exponent $p$ of the gap-decay curve as a direct estimate of R1's predicted rate ($p \ge 2$ for regression). We caution against this. The fitted exponent and the population rate diverge for the reasons below.

R1 is a statement about \emph{population} quantities: the excess risk of the best linear predictor over the Bayes-optimal predictor. As a positive sum of terms $\prod_j \rho_j^{\alpha_j}$ with $|\alpha| \ge 2$, this population excess risk has an effective single-power exponent of at least 2 over any range of $\sigma$, and, for a target with degree-2 through degree-4 content like our synthetic benchmark, about 2.9. The empirical quantity we actually observe, however, is the difference of two \emph{finite-sample} test scores (kernel ridge minus ridge), which is not the same object. Three controlled observations show how the two come apart:

\begin{itemize}
\item \textbf{1D benchmark (known ground truth).} On the synthetic 1D benchmark, the population excess risk has fitted exponent $2.9$. The fitted exponent of the realized estimator gap, by contrast, ranges from about $3.2$ to $4.6$ across training sizes $n \in \{200, \ldots, 3200\}$, \emph{above} the population value, because at high noise the small attenuated gap is swamped by estimation variance and the nonlinear model fails to beat ridge sooner than the population gap closes.
\item \textbf{Multivariate synthetic.} In a Gaussian feature model with linear and pairwise-interaction structure, where the gap remains positive (low dimensionality, $d \le 5$), the realized exponent matches the population value ($\approx 2.0$--$2.2$ against a population value of $2.0$). At higher dimensionality ($d \ge 20$, $n = 10{,}000$) the nonlinear model cannot capture the interactions and fails to beat ridge even at $\sigma = 0$, the statistical-suppression regime, so no positive gap, and no exponent, exists to measure. In neither regime does the controlled realized exponent fall below 2.
\item \textbf{UK Biobank tasks.} On real features, by contrast, the fitted exponent does fall below 2 (median $1.7$) and declines with feature dimensionality (Spearman $\rho = -0.34$; median $1.86$ for $d \le 31$ versus $1.48$ for $d \ge 45$). The population identity contains no such dimensionality dependence, and our controlled experiments never produce sub-2 exponents. The real-data values therefore reflect departures from the idealized model (non-Gaussian feature marginals, feature correlation, pre-existing measurement noise, and nonlinear structure that is not a clean low-degree Hermite expansion) rather than a property of $E[Y \mid \tilde{X}]$. We do not attempt to attribute the sub-2 exponents to a single one of these.
\end{itemize}

The fitted exponent is therefore not a reliable estimate of R1's population rate: in controlled settings it matches or exceeds the rate, and on real data it is pulled below it by departures from the model's assumptions. The robust, theory-consistent readouts on real data are instead the \emph{direction} of the effect (the nonlinear gap is preferentially destroyed, decay ratio $> 1$) and its \emph{reversibility} (averaging replicates restores the gap). The quantitative rate claim is carried by the controlled benchmark, where the population excess risk is computable, not by the biobank data.

\subsection{Domain table: ICC comparability}

The ICC values in Table~\ref{tab:domains} of the main text come from published test-retest studies with varying designs: scan-rescan intervals range from minutes (same-session rescans for MRI) to weeks (Olink proteomics). ICC values from shorter intervals tend to be higher because they exclude slow biological drift. When comparing ICC across modalities, this heterogeneity should be kept in mind. The values reported are representative of the field's published ranges, not from a single standardized protocol.

\section{Synthetic benchmark specifications}
\label{sec:si-synthetic}

All synthetic experiments use Gaussian features $X \sim N(0, 1)$ (or $N(0, I_d)$ in the multivariate case) with independent additive Gaussian noise unless otherwise noted. Hyperparameters are selected on a held-out validation set.

\subsection{1D regression benchmark}

\textbf{Target function.} $f(z) = 0.90 \, \mathrm{He}_1(z) - 0.55 \, \mathrm{He}_2(z) + 0.35 \, \mathrm{He}_3(z) + 0.18 \, \mathrm{He}_4(z)$, where $\mathrm{He}_k$ are probabilist Hermite polynomials. This gives a known Hermite spectrum with energy at degrees 1--4. Its population excess risk has effective decay exponent $\approx 2.9$ (used as the ground-truth reference in \S\ref{sec:si-proxy}).

\textbf{Data generation.} $Y = f(Z) + \varepsilon$, $\varepsilon \sim N(0, 0.45^2)$, $Z \sim N(0,1)$. Observed features: $\tilde{X} = Z + \eta$, $\eta \sim N(0, \sigma^2)$.

\textbf{Sample sizes.} $n_{\mathrm{train}} = 400$, $n_{\mathrm{val}} = 200$, $n_{\mathrm{test}} = 2{,}500$. Repeated across 12 independent splits.

\textbf{Injection arm (Fig.~\ref{fig:synthetic}a-b).} Noise levels $\sigma \in \{0, 0.2, 0.5, 0.75, 1.0, 1.5, 2.0\}$.

\textbf{Reduction arm (Fig.~\ref{fig:synthetic}c-d).} Baseline noise $\sigma_0 = 2.0$. Replicate counts $m \in \{1, 2, 4, 8, 16\}$. Replicates are averaged before fitting: $\bar{X} = \frac{1}{m}\sum_{i=1}^m \tilde{X}^{(i)}$.

\textbf{Estimators.} ridge regression ($\alpha \in \{10^{-4}, \ldots, 10^1\}$) versus degree-4 polynomial kernel ridge ($\alpha \in \{10^{-5}, \ldots, 10^1\}$, $\gamma \in \{0.25, 0.5, 1.0, 2.0\}$, $\mathrm{coef}_0 = 1$). Fig.~S1 adds XGBoost (100 trees, max depth 4, learning rate 0.1) and MLP (2 hidden layers, 128 units, ReLU, Adam optimizer, 200 epochs).

\textbf{Metric.} $R^2$ on the held-out test set.

\subsection{Fig.~S3: MNIST classification}

\textbf{Dataset.} MNIST handwritten digits (10 classes, $28\times28$ pixels). Reduced to 20 whitened PCA components (fitted on training data only).

\textbf{Noise injection.} Additive Gaussian noise in whitened PCA space at $\sigma \in \{0, 0.25, 0.5, 0.75, 1.0, 1.25, 1.5\}$.

\textbf{Reduction arm.} Baseline noise $\sigma_0 = 1.5$, replicate counts $m \in \{1, 2, 4, 8, 16\}$.

\textbf{Sample sizes.} $n_{\mathrm{train}} = 2{,}000$, held-out validation and test. 5 repeats.

\textbf{Estimators.} Logistic regression (L2-penalized) versus RBF kernel ridge classifier.

\subsection{Fig.~S4: Non-Gaussian noise robustness}

\textbf{Target function.} $f(z) = 0.8 \, \mathrm{He}_1(z) + 0.6 \, \mathrm{He}_3(z)$, giving $k_{\min} = 3$ and analytical excess risk $= 0.6^2 \cdot \rho^3$.

\textbf{Noise distributions.} All matched to variance $\sigma^2$:
\begin{itemize}
\item Gaussian: $\eta \sim N(0, \sigma^2)$
\item Laplace: $\eta \sim \mathrm{Laplace}(0, \sigma/\sqrt{2})$
\item Uniform: $\eta \sim \mathrm{Uniform}(-\sqrt{3}\sigma, \sqrt{3}\sigma)$
\item Gaussian mixture: $\eta \sim 0.9 \, N(0, \sigma^2/c) + 0.1 \, N(0, 10\sigma^2/c)$ with $c = 0.9 + 1.0$
\end{itemize}

\textbf{Sample sizes.} $n_{\mathrm{train}} = 2{,}000$, $n_{\mathrm{test}} = 5{,}000$. 10 repeats. Label noise $\varepsilon \sim N(0, 0.2^2)$.

\textbf{Noise levels.} $\sigma \in \{0, 0.25, 0.5, 0.75, 1.0, 1.5\}$.

\subsection{Fig.~S5: Selective linearization}

\textbf{Features.} $d = 10$, split into Reliable ($j = 0, \ldots, 4$, $\sigma_j = 0.5t$) and Unreliable ($j = 5, \ldots, 9$, $\sigma_j = 1.5t$) tiers, swept over $t \in \{0, 0.25, 0.5, 0.75, 1.0, 1.5, 2.0, 3.0\}$. Features are uncorrelated, so the per-feature reliability product of R1 applies exactly (\S\ref{sec:si-r1}).

\textbf{Target function.} Degree-2 Hermite terms only (pairwise interactions), partitioned into three classes: RR (reliable-reliable, 10 pairs), RU (reliable-unreliable, 25 pairs), UU (unreliable-unreliable, 10 pairs). Each class carries equal total $L^2$ energy at $t = 0$ (energy $= 1.0$ per class). Coefficients are random with controlled norm.

\textbf{Sample sizes.} $n_{\mathrm{train}} = 2{,}000$, $n_{\mathrm{test}} = 5{,}000$. 8 repeats. Label noise $\varepsilon \sim N(0, 0.3^2)$.

\subsection{Fig.~S7: Noise-sample-size interaction}

\textbf{Same 1D benchmark} as in main Fig.~\ref{fig:synthetic}, with the following grid:

\textbf{Sample sizes.} $n \in \{100, 200, 400, 800, 1{,}600, 3{,}200, 6{,}400\}$.

\textbf{Noise levels.} $\sigma \in \{0, 0.1, 0.2, 0.3, 0.5, 0.75, 1.0, 1.25, 1.5, 2.0, 2.5, 3.0, 3.5, 4.0\}$.

Each $(n, \sigma)$ cell is repeated across 12 splits. Panel (b) shows the marginal gain $\Delta R^2(n{=}6400) - \Delta R^2(n{=}100)$ as a function of $\sigma$.

\section{Lean 4 formalization}
\label{sec:si-lean}

Both R1 and C1 are formalized and machine-verified in the Lean~4 proof assistant using Mathlib and 12 additional library modules that formalize results not yet available in Mathlib (Mehler's formula, Hermite orthogonality, Pinsker's inequality, Gaussian KL divergence, and others). No custom axioms beyond Lean's standard foundational axioms (\texttt{propext}, \texttt{Quot.sound}, \texttt{Classical.choice}) are used. The minimal proof subset accompanying this paper comprises 22 files (12 library + 10 theorem modules) with 0 sorries. We note that the $\sigma^4$ sample-complexity scaling (\S\ref{sec:si-sigma4}) is deliberately \emph{not} part of the formalization, as it is a heuristic rather than a theorem.

\subsection{Regression (R1)}

\begin{itemize}
\item \texttt{RegressionSetup.lean}: posterior distribution, best linear predictor definition.
\item \texttt{HermiteShrinkage.lean}: Hermite coefficient shrinkage, Parseval identity, exact excess MSE formula. Key theorem: \texttt{excess\_mse\_hermite}.
\item \texttt{RegressionHermiteRate.lean}: concrete excess MSE as $\sum_{k \ge 2} (a_k \cdot \rho^k)^2 \cdot k!$ (1D), rate bound $O(\rho^4)$.
\end{itemize}

\subsection{Classification (C1)}

The coupling bound is proved in five files: \texttt{CouplingBound.lean} $\to$ \texttt{MixtureGaussian.lean} $\to$ \texttt{GaussianEnvelope.lean} $\to$ \texttt{DiagonalGaussian.lean} $\to$ \texttt{Theorem2Assembly.lean}. Two additional files (\texttt{ScalarGaussian.lean}, \texttt{ExcessRiskIntegral.lean}) provide shared definitions used by both the regression and classification chains.

\subsection{Library modules (12 files)}

Key modules: \texttt{MehlerFormula.lean} (Mehler's formula), \texttt{HermiteOrthogonality.lean} ($\int \mathrm{He}_m \mathrm{He}_n \varphi = n! \delta_{mn}$), \texttt{PinskerScalar.lean} and \texttt{PinskerDensity.lean} (Pinsker's inequality), \texttt{KLGaussScalar.lean} and \texttt{GaussianKL.lean} (Gaussian KL divergence), \texttt{SteinLemma.lean} (Stein's identity).

\subsection{Verification}

\begin{verbatim}
cd lean-proof-minimal && lake build
\end{verbatim}

\noindent All 22 files compile against Lean~4's kernel with 0 sorries (proof gaps).

\section*{Supplementary Figures}
\addcontentsline{toc}{section}{Supplementary Figures}

\begin{figure}[htbp]
\centering
\includegraphics[width=\textwidth]{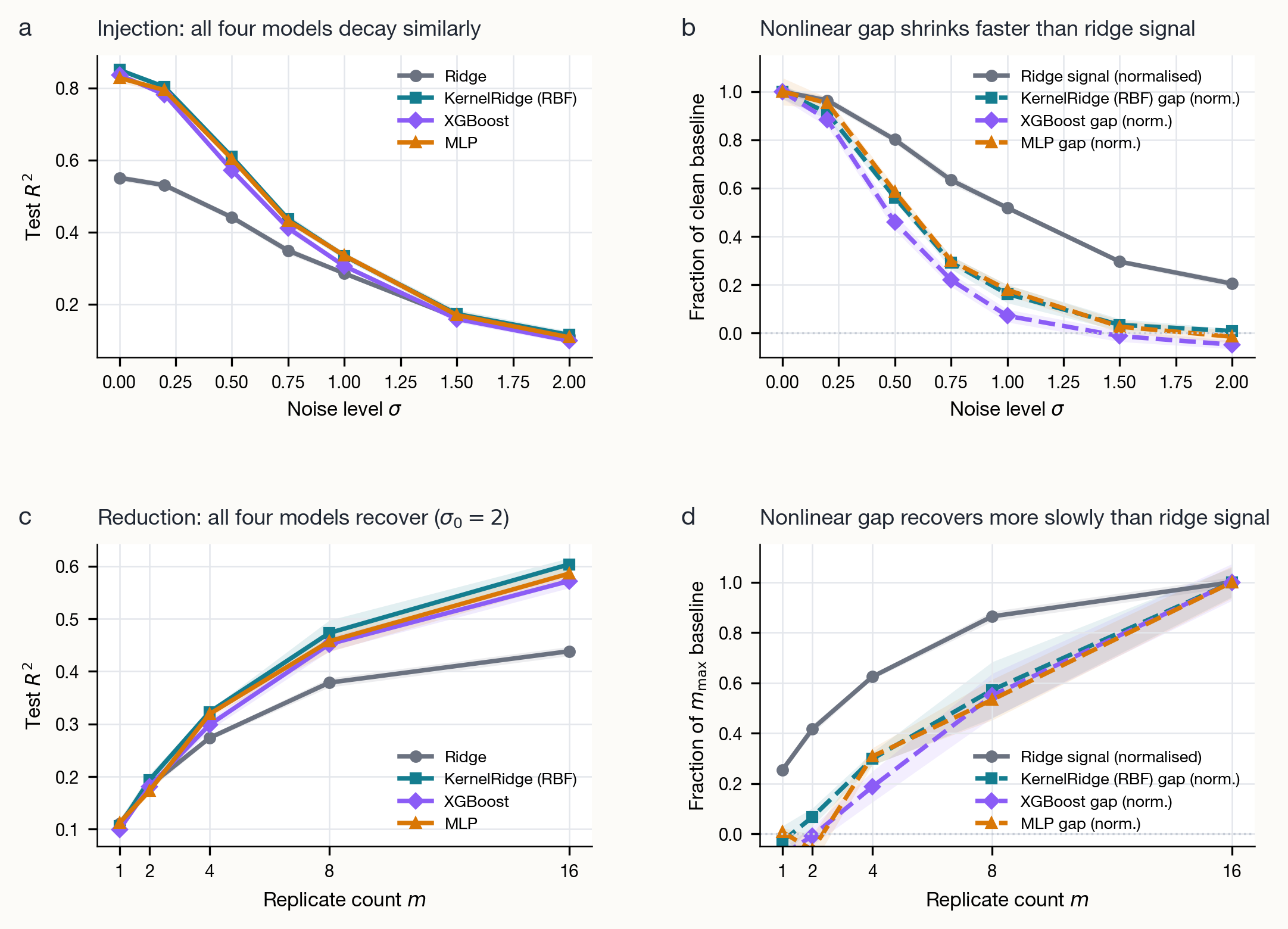}
\caption{\textbf{The rate-mismatch pattern holds across nonlinear model classes.} The synthetic benchmark from Fig.~\ref{fig:synthetic} repeated with four models: ridge, RBF kernel ridge, XGBoost, and MLP (2-layer, 128 units). (a)~Injection: all four models decay under noise. (b)~Normalized: the nonlinear gaps (kernel ridge, XGBoost, MLP minus ridge) decay faster than the ridge signal for all three nonlinear models. (c--d)~Reduction: averaging replicates recovers performance, with all three nonlinear models recovering faster than ridge.}
\label{fig:model-class-audit}
\end{figure}

\begin{figure}[htbp]
\centering
\includegraphics[width=\textwidth]{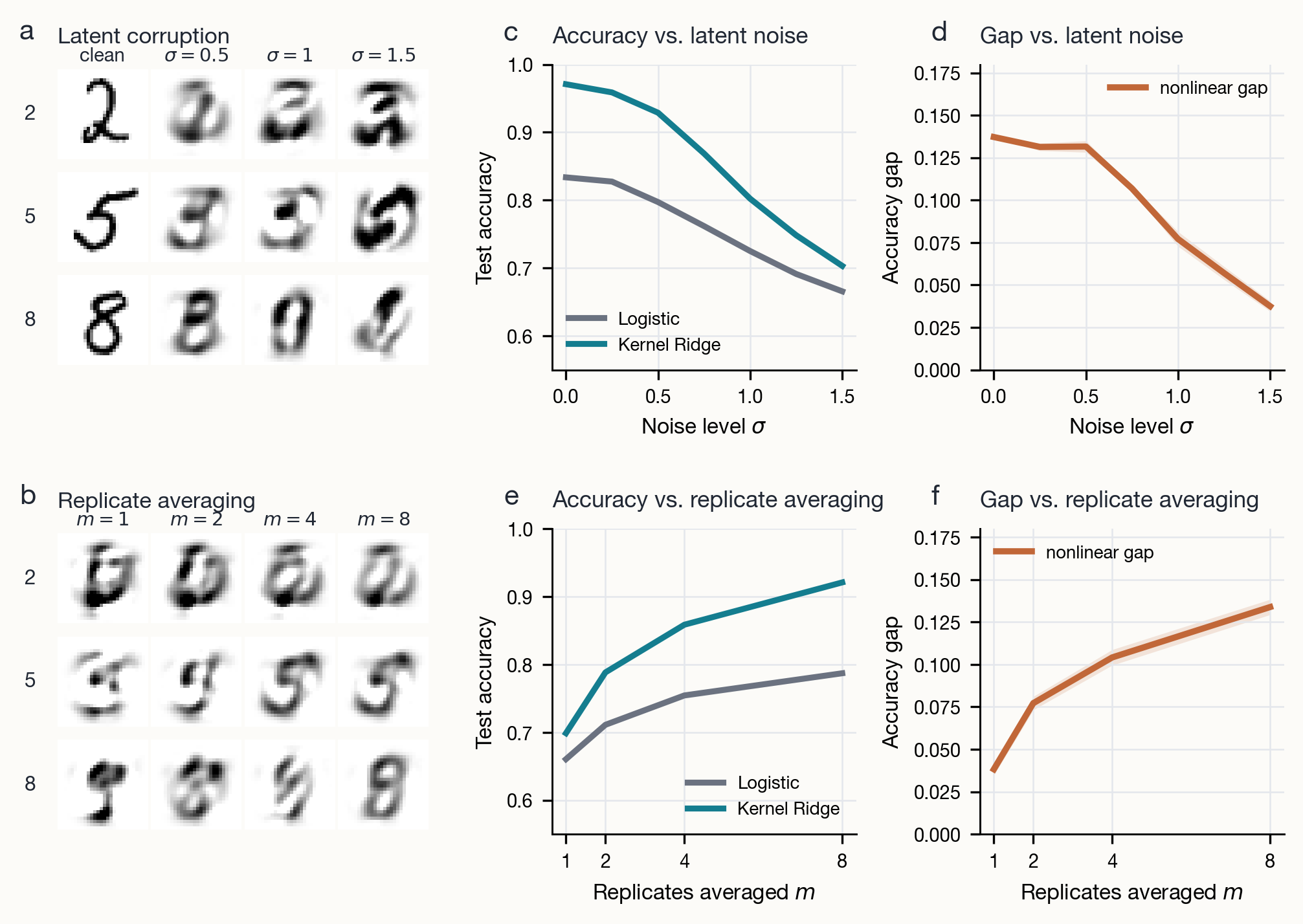}
\caption{\textbf{Rate mismatch on MNIST digit classification.} Logistic regression versus RBF-kernel SVM on MNIST digits with injected pixel noise (left) and replicate averaging (right). The classification gap decays under noise injection and recovers under averaging, paralleling the regression result in Fig.~\ref{fig:synthetic}.}
\label{fig:mnist}
\end{figure}

\begin{figure}[htbp]
\centering
\includegraphics[width=\textwidth]{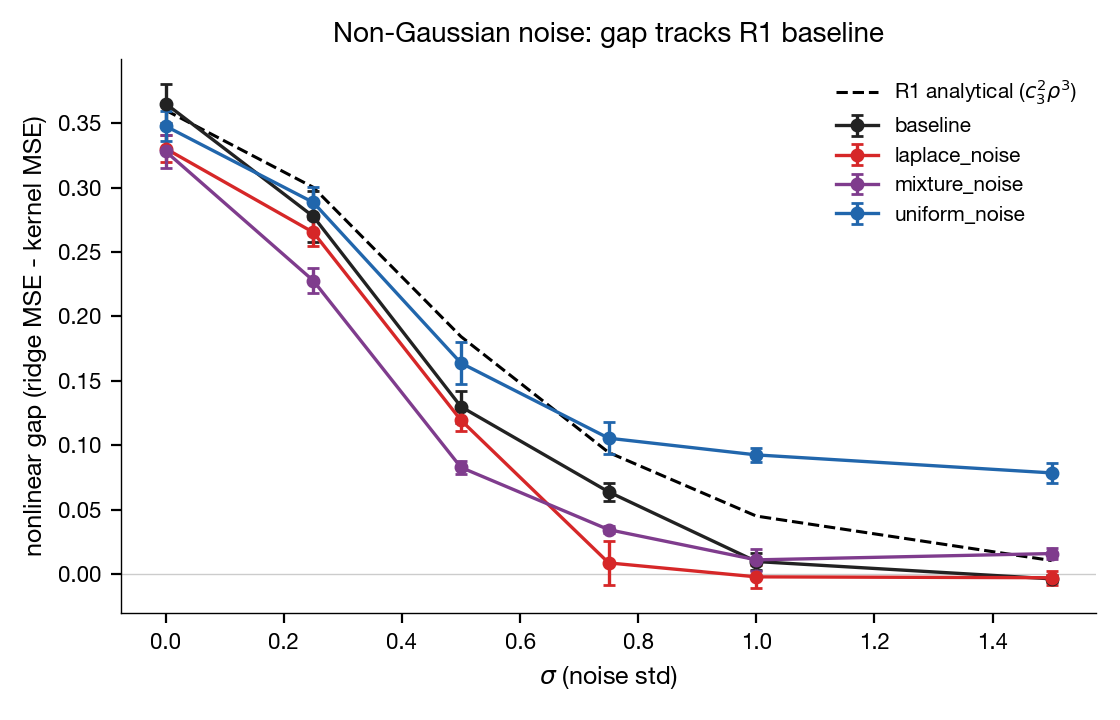}
\caption{\textbf{Robustness to non-Gaussian noise distributions.} The synthetic benchmark repeated with four noise distributions: Gaussian, Laplace (heavy-tailed), uniform (light-tailed), and a Gaussian mixture (bimodal). The rate-mismatch pattern (faster decay of the nonlinear gap) is qualitatively preserved across all distributions, consistent with the argument that the product attenuation structure holds for any finite-variance noise.}
\label{fig:robustness}
\end{figure}

\begin{figure}[htbp]
\centering
\includegraphics[width=\textwidth]{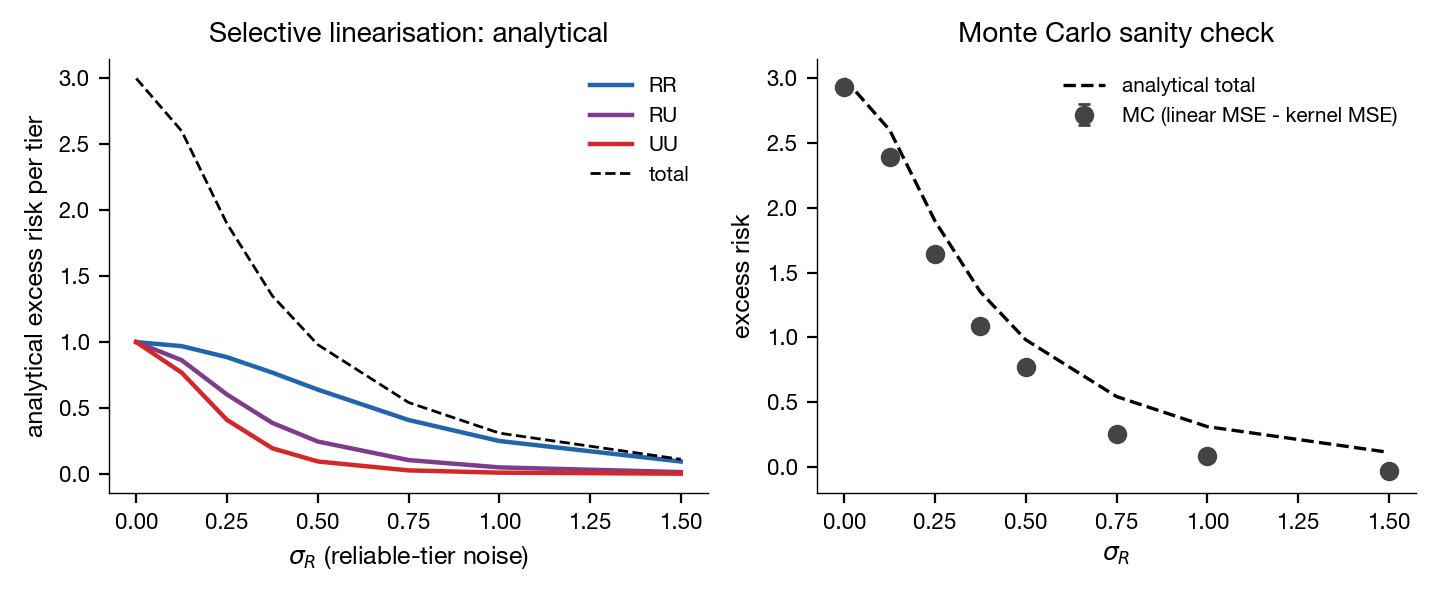}
\caption{\textbf{Selective linearization by per-feature reliability.} (Left)~Analytical excess risk decomposed by interaction type: RR (reliable-reliable, blue), RU (reliable-unreliable, purple), UU (unreliable-unreliable, red), and total (dashed). Interactions involving unreliable features decay fastest, while reliable-reliable interactions are preserved, consistent with R1's product structure. (Right)~Monte Carlo validation: empirical excess risk (linear MSE minus kernel MSE, dots) tracks the analytical prediction (dashed).}
\label{fig:selective}
\end{figure}

\begin{figure}[htbp]
\centering
\includegraphics[width=\textwidth]{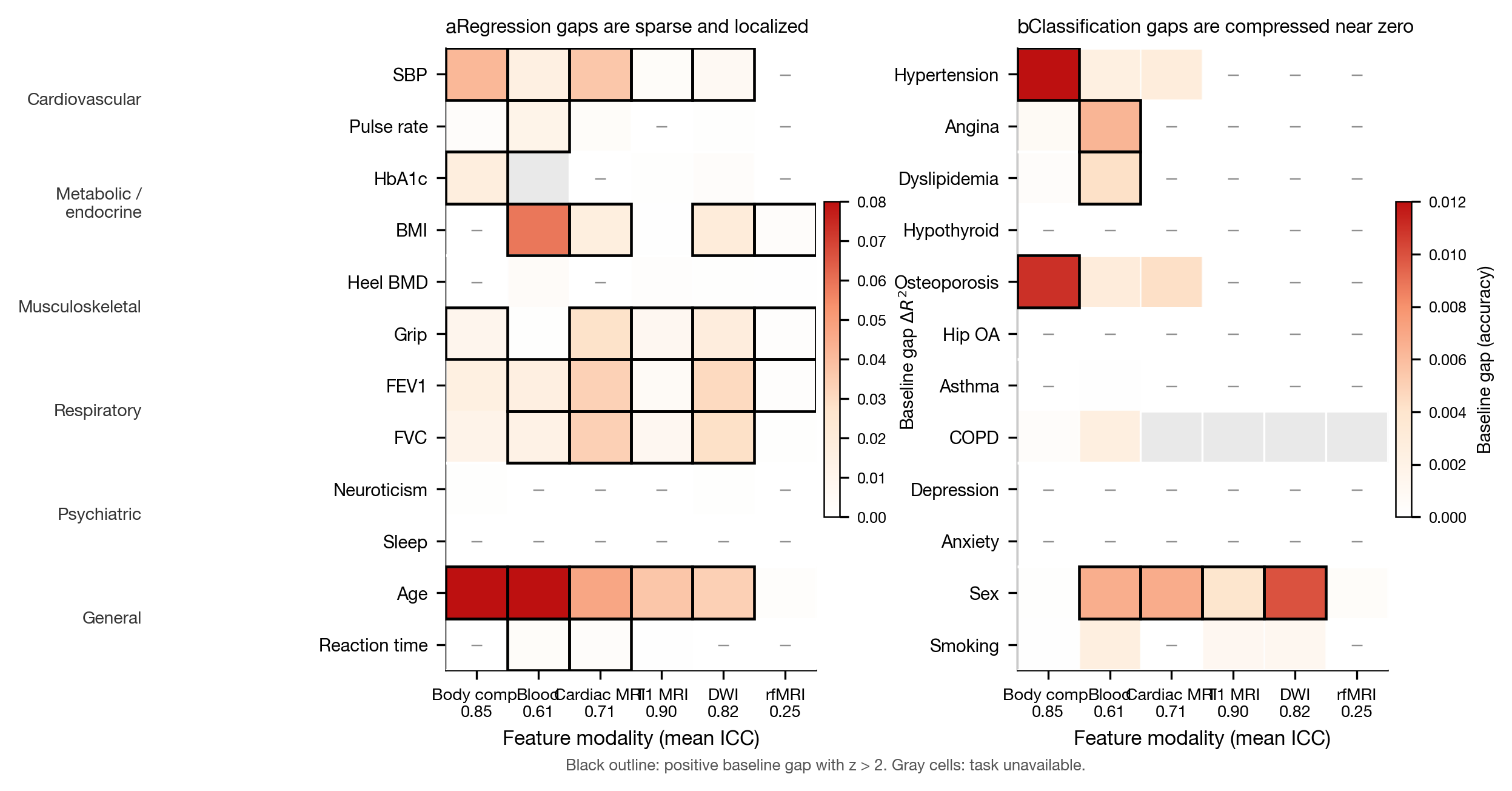}
\caption{\textbf{Baseline nonlinear gap across 140 UK Biobank tasks.} (a)~Regression tasks. (b)~Classification tasks. Each cell shows the baseline nonlinear gap (kernel ridge minus ridge score at $\sigma^2 = 0$) for one target (row) predicted from one feature modality (column). Black outlines indicate statistically significant gaps ($z > 2$). Grey cells indicate unavailable task-modality combinations. Modalities are ordered by published mean ICC (left to right: body composition 0.85, blood 0.61, cardiac MRI 0.71, T1 MRI 0.90, DWI 0.82, rfMRI 0.25). Regression gaps are sparse and concentrated in high-ICC modalities; classification gaps are compressed near zero.}
\label{fig:landscape}
\end{figure}

\begin{figure}[htbp]
\centering
\includegraphics[width=\textwidth]{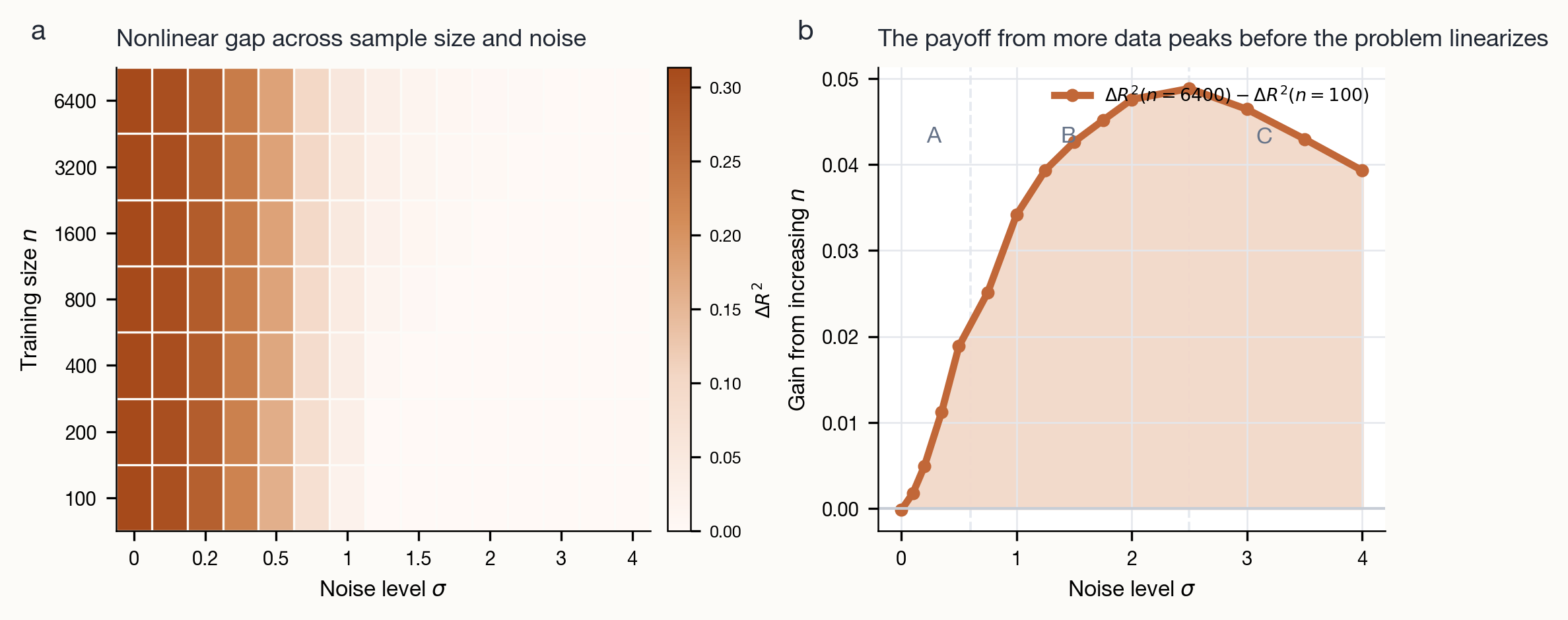}
\caption{\textbf{The compounding effect of noise on sample requirements.} (a)~Nonlinear gap $\Delta R^2$ (color) as a function of training set size $n$ (vertical) and injected noise $\sigma$ (horizontal) on the synthetic benchmark from Fig.~\ref{fig:synthetic}. The gap is largest at high $n$ and low $\sigma$ (upper left) and vanishes as noise increases or sample size shrinks. (b)~The marginal payoff from increasing $n$ (from 100 to 6,400) as a function of noise. The payoff peaks at intermediate noise and declines at high noise, where the problem has been linearized and additional data cannot recover the nonlinear signal.}
\label{fig:scissors}
\end{figure}

\end{document}